\documentclass[sigconf]{acmart}
\usepackage[shortlabels]{enumitem}
\usepackage{multirow}
\usepackage{times}
\usepackage{subfig}
\usepackage[linewidth=1pt]{mdframed}
\usepackage[export]{adjustbox}

\usepackage{array}
\newcolumntype{L}[1]{>{\raggedright\let\newline\\\arraybackslash\hspace{0pt}}m{#1}}
\newcolumntype{C}[1]{>{\centering\let\newline\\\arraybackslash\hspace{0pt}}m{#1}}
\newcolumntype{R}[1]{>{\raggedleft\let\newline\\\arraybackslash\hspace{0pt}}m{#1}}

\usepackage[linesnumbered,ruled,vlined]{algorithm2e}

\newcommand{\system}{{\sc ExplainIt}}
\newcommand{\hotel}{\mbox{\sc Hotel}}
\newcommand{\restaurant}{\mbox{\sc Restaurant}}
\newcommand{\yelp}{\mbox{\sc Yelp}}
\newcommand{\op}[1]{{``\textsl{#1}''}}

\newcommand{\asop}[1]{{\textsl{#1}}}
\newcommand{\expcls}{{MaskedDualAttn}}
\newcommand{\canonical}{{WS-OPE}}

\setlist[itemize]{leftmargin=*}
\setlist[enumerate]{leftmargin=*}

\AtBeginDocument{%
  \providecommand\BibTeX{{%
    \normalfont B\kern-0.5em{\scshape i\kern-0.25em b}\kern-0.8em\TeX}}}

\copyrightyear{2021}
\acmYear{2021}
\setcopyright{iw3c2w3}
\acmConference[WWW '21]{Proceedings of the Web Conference 2021}{April 19--23, 2021}{Ljubljana, Slovenia}
\acmBooktitle{Proceedings of the Web Conference 2021 (WWW '21), April 19--23, 2021, Ljubljana, Slovenia}
\acmPrice{}
\acmDOI{10.1145/3442381.3450081}
\acmISBN{978-1-4503-8312-7/21/04}
\begin{document}
\title{Constructing Explainable Opinion Graphs from Reviews}
\author{Nofar Carmeli}
\authornote{Work done during internship at Megagon Labs.}
\affiliation{%
  \institution{Technion}
  \city{Haifa}
  \country{Israel}
}
\email{snofca@cs.technion.ac.il}

\author{Xiaolan Wang}
\affiliation{%
  \institution{Megagon Labs}
  \city{Mountain View}
  \country{USA}
}
\email{xiaolan@megagon.ai}

\author{Yoshihiko Suhara}
\affiliation{%
  \institution{Megagon Labs}
  \city{Mountain View}
  \country{USA}
}
\email{yoshi@megagon.ai}

\author{Stefanos Angelidis}
\affiliation{%
  \institution{University of Edinburgh}
  \city{Edinburgh}
  \country{UK}
}
\email{s.angelidis@ed.ac.uk}

\author{Yuliang Li}
\affiliation{%
  \institution{Megagon Labs}
  \city{Mountain View}
  \country{USA}
}
\email{yuliang@megagon.ai}

\author{Jinfeng Li}
\affiliation{%
  \institution{Megagon Labs}
  \city{Mountain View}
  \country{USA}
}
\email{jinfeng@megagon.ai}

\author{Wang-Chiew Tan}
\authornote{Work done while at Megagon Labs.}
\affiliation{%
  \institution{Facebook AI}
  \city{Menlo Park}
  \country{USA}
}
\email{wangchiew@fb.com}

\renewcommand{\shortauthors}{Carmeli, N., Wang, X., Suhara, Y., Angelidis, S., Li, Y., Li, J., and Tan, W.C.}

\begin{abstract}
The Web is a major resource of both factual and subjective information. While there are significant efforts to organize factual information into knowledge bases, there is much less work on organizing opinions, which are abundant in subjective data, into a structured format.

We present \system, a system that extracts and organizes opinions into an opinion graph, which are useful for downstream applications such as generating explainable review summaries and facilitating search over opinion phrases. In such graphs, a node represents a set of semantically similar opinions extracted from reviews and an edge between two nodes signifies that one node explains the other. \system{} mines explanations in a supervised method and groups similar opinions together in a weakly supervised way before combining the clusters of opinions together with their explanation relationships into an opinion graph. We experimentally demonstrate that the explanation relationships generated in the opinion graph are of good quality and our labeled datasets for explanation mining and grouping opinions are publicly available at \url{https://github.com/megagonlabs/explainit}.
\end{abstract}

\ccsdesc[500]{Information systems~Web mining}

\keywords{Opinion mining, explanation, opinion graph construction}

\maketitle

\section{Introduction}\label{sec:intro}
The Web is a major resource for people to acquire information, whether factual or subjective. In recent years, there have been significant advances in extracting facts in the form of subject-predicate-object triples and constructing knowledge bases of such facts~\cite{weikum2010information, dong2014knowledge, nickel2015review, mitchell2018never}. In comparison, there are much less efforts around 
constructing organized knowledge bases of opinions~\cite{bhutanisampo}, which are abundant in subjective data, such as reviews and tweets. 
In fact, according to a recent study\footnote{\url{https://fanandfuel.com/no-online-customer-reviews-means-big-problems-2017/}}, more than $90\%$ of customers read reviews before committing on visiting a business or making a purchase. A natural question is thus the following: is there a systematic way to organize opinions into knowledge bases that will make it easier for customers to understand the opinions found in subjective data?

Existing opinion mining techniques~\cite{hu2004aaai, qiu2011coling, liu2012sentiment, pontiki2015semeval, pontiki2016semeval, xu2019bert} cannot be directly applied to organize
the extracted opinions. First, they largely focus on improving the 
accuracy of opinion extraction and aspect-based sentiment analysis of the extracted opinions over a set of predefined aspects. They cannot be used, in particular, to determine the relationships between opinions. For example, while they can determine the sentiment of an extracted opinion \op{very good location}, they cannot explain why the location is very good in relation to other extracted opinions.
Furthermore, simply collecting all extracted opinions will result in a lot of redundancy and may also lead to incorrect conclusions. 
For example, if the list of all extracted opinions are \{\op{quiet room}, \op{very noisy street}, \op{loud neighborhood}, \op{horrible city noise}, \op{quiet room}\}, one can incorrectly conclude that \op{quiet room} is the most popular opinion if the opinions are not organized according to similarity. 
An early attempt~\cite{bhutanisampo} that produces knowledge bases of opinions does not fully address the above limitations since it does not de-duplicate similar opinions, nor considers the direction of explanation between opinions. 

\begin{figure*}[t]
\begin{minipage}[t][][b]{0.45\textwidth}
\centering
\begin{mdframed}
\small
\begin{tabular}{l}
\textbf{Review \#1, Score 85, Date: Feb. 30, 2020}  \\
\multicolumn{1}{p{7cm}} {\it Pros: Friendly and helpful staff. Great location. Walking distance to Muni bus stops. Not too far away from Fisherman's Wharf, Aquatic Park. Cons: extremely noisy room with paper thin walls.}  \\ \bottomrule
\textbf{Review \#2, Score 80, Date: Feb. 30, 2019}  \\
\multicolumn{1}{p{7cm}} {\it Good location close to the wharf, aquatic park and the many other attractions. the rooms are ok but a bit noisy, loud fridge and AC.
}  \\
\end{tabular}
\end{mdframed}
\vspace{-2mm}
\end{minipage}
\begin{minipage}[t][][b]{0.5\textwidth}
\vspace{2mm}
\includegraphics[width=\linewidth]{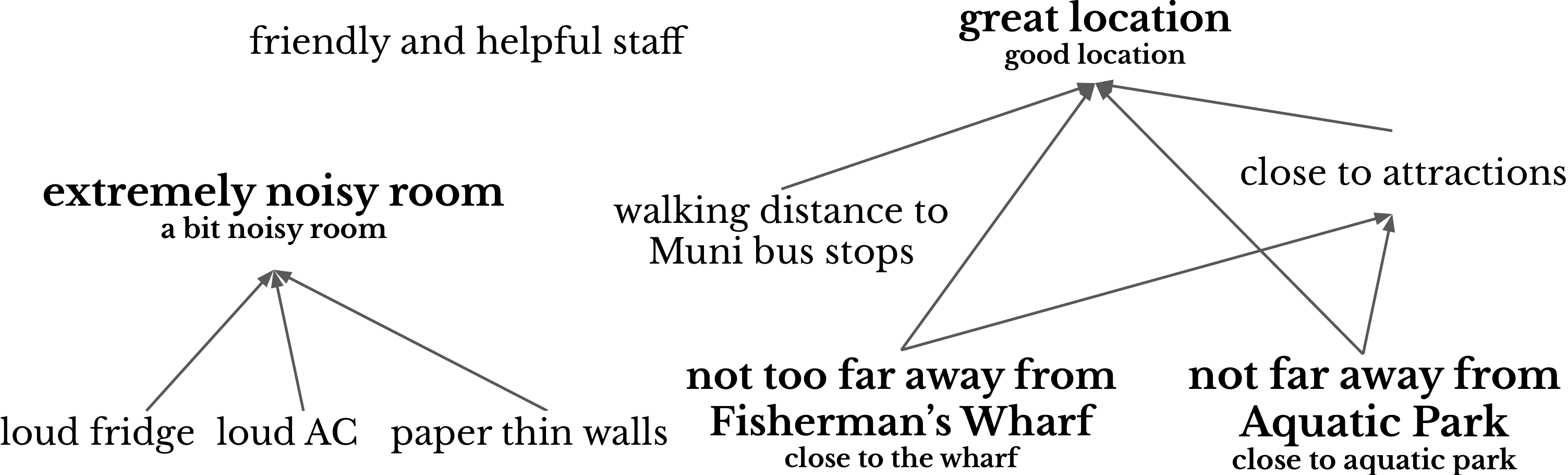}
\end{minipage}
    \caption{Opinion Graph (right) based on opinions extracted from reviews (left). A node explains its parent node.}
\label{fig:example}
\end{figure*}

With the above observations, we asked ourselves the following question:
\textit{Can we go beyond opinion mining to represent opinions and the relationships among them uniformly into a knowledge base?}
To understand how best to organize opinions into a knowledge base, we analyzed the properties of subjective information in reviews through a series of annotation tasks and confirmed that:
\begin{itemize}
    \item {\em Opinions phrases}, or opinions in short, are pairs of the form (opinion term, aspect term) such as (\op{very good}, \op{location}). Opinions are the most common expression for subjective information in reviews. In $100$ random review sentences that we annotated, we observed that $84.75\%$ of subjective information is in this form. 
    \item {\em Explanation}, or inference, is the most common relationship between opinions that are correlated in reviews.
    We annotated opinions that co-occur in $40$K random review sentences\footnote{Under the Appen platform (\url{https://appen.com/}).} and observed that $12.3\%$ of the opinions are correlated under some relationship (e.g., one explains/contradicts/paraphrases the other). Among these opinions, $74.2\%$ of the opinions are related under the explanation relationship, which is the focus of this paper.
    \item Many opinions and relationships between opinions are oriented around specific entities, not across multiple entities.
    For example, the opinion \op{close to main street} explains \op{very noisy room} for a specific hotel in the review ``Our room was very noisy as it is close to the main street''. However, this explanation may not be true for arbitrary hotels.
\end{itemize}

Based on this analysis, we propose a graph representation for organizing opinions, called the \textsl{Opinion Graph} that organizes opinions around the explanation relationship based on reviews specific to an entity.
A node is an opinion of
the form (opinion term, aspect term) and consists of all opinions that are close to the node according to their semantic similarity. An edge ($u$,$v$) between two nodes $u$ and $v$ denotes that $u$ explains $v$. 
We found this to be a versatile structure for organizing opinions of reviews because (a) the opinion graph is a concise and structured representation of the opinions over lots of reviews, 
(b) the nodes can aggregate and represent opinions at different levels of granularity, (c) the edges explain the opinions based on other opinions that appear in the reviews, (d) the provenance of opinions in nodes can be traced back to the input reviews where they are extracted from, and (e) the opinion graph is a useful abstraction that supports a series of downstream applications, from the generation of explainable review summaries to facilitating search over opinion phrases or criteria~\cite{Li:2019:Opine}. 

The right of Figure~\ref{fig:example} shows an opinion graph that is generated from the hotel reviews on the left of the figure. Each node in the graph represents a set of semantically similar opinions. Each opinion consists of an opinion term, followed by an aspect term. For example, \op{good location}, where \textsl{``good''} is an opinion term and \textsl{``location''} is an aspect term. Each edge represents the explanation relationship between opinions. For example, \op{paper thin walls} explains \op{extremely noisy room}.
This opinion graph enables one to easily create a customized summary of the reviews by using the entire graph, or only for portions of the graph, such as which attractions the hotel is in close proximity with. 
Moreover, end users or downstream applications can navigate aspects and opinions based on their specific needs and seek explanations of the extracted opinions, e.g., understand why the hotel is \textsl{``extremely noisy''} or why it is in a \textsl{``great location''}. A prototype based on this application has been demonstrated~\cite{wang2020extremereader}.

Opinion Graphs are constructed from reviews through a novel opinion graph construction pipeline \system, which we will present in this paper. To the best of our knowledge, \system\ is the first pipeline that can extract and organize both opinions and their explanation relationships from reviews. 
It is challenging to construct an opinion graph from reviews. First, the review sentences are inherently noisy and can be nuanced. Hence, mining
opinions and the explanation relationships between them can be difficult. Second, all opinions and their predicted explanation relationships need to be integrated into one opinion graph, while taking into account potential inaccuracies and the noise and nuances inherent in languages.
To summarize, we make the following contributions:

\begin{itemize} 
    \item We develop \system, a system that generates an opinion graph about an entity from a set of reviews about the entity. \system{} (a) mines opinion phrases, (b) determines the explanation relationships between them, (c) canonicalizes semantically similar opinions into opinion clusters, and (d) generates an opinion graph for the entity from the inferred explanation relationships and opinion clusters. In particular, our technical contributions include the explanation classifier, the in-domain training data, and the opinion phrase learning mechanism for opinion canonicalization.
    \item We evaluate the performance of \system\ through a series of experiments. We show that our explanation classifier performs $5\%$ better than a fine-tuned BERT model~\cite{devlin2018bert} and $10\%$ better than a re-trained textual entailment model~\cite{parikh-etal-2016-decomposable}. We show that learned opinion phrase representations are able to improve existing clustering algorithms by up to $12.7\%$ in V-measure.  
    Finally, our user study shows that human judges agree with the predicted graph edges produced by our system in more than $77\%$ of the cases. 
    \item Our crowdsourced labeled datasets (in the hotel and restaurant domains) for two subtasks (mining explanation relationships and canonicalizing semantically similar opinion phrases) that we use for training and evaluation are publicly available at \url{https://github.com/megagonlabs/explainit}.
\end{itemize}

\noindent
{\bf Outline~}
We give an overview of \system\ in Section~\ref{sec:overview}. We present the component for mining explanation relationships in Section~\ref{sec:exp}. We demonstrate how we canonicalize similar opinion phrases in Section~\ref{sec:cluster} and how we construct an opinion graph in Section~\ref{sec:graph}. We evaluate \system\ in Section~\ref{sec:eval}. We outline related work in Section~\ref{sec:related} and conclude this paper in Section~\ref{sec:conclusion}.

\section{Preliminaries} \label{sec:overview}
\begin{figure}[t]
    \centering
    \includegraphics[width=0.48\textwidth]{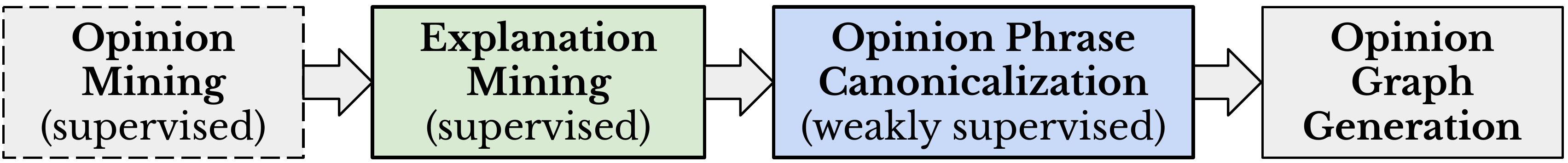}
    \caption{\system{} pipeline. The explanation Mining module and the Opinion Phrase Canonicalization module form our major technical contributions.}
    \label{fig:overview}
    \vspace{-2mm}
\end{figure}

An \textsl{opinion phrase} is a pair $p=(o, a)$, where $o$ is the \textsl{opinion term}, and $a$ is the aspect term $o$ is referring to. 
For example, the last sentence of Review \#1 in Figure~\ref{fig:example}, ``Cons: extremely noisy room with paper thin walls'', contains two opinion phrases: $p_1=$ (``\asop{extremely noisy}'', ``\asop{room}''); and $p_2=$ (``\asop{paper thin}'', ``\asop{walls}''). 
An \textsl{explanation} $e = (p_i\rightarrow p_j)$ is a relationship between two opinion phrases, where $p_i$ explains $p_j$. For example, (\asop{``paper thin walls''} $\rightarrow$ \asop{``extremely noisy room''}) and (\asop{``close to attractions''} $\rightarrow$ \asop{``great location''}) are two valid explanations.\smallskip

\noindent \textsc{\textbf{Definition:}} An \textsl{Opinion  Graph} $G = (N,E)$ for a set $S$ of opinion phrases is such that 
(1) every opinion phrase $p\in S$ belongs to exactly one node $n\in N$,
(2) each node $n\in N$ consists of {\em semantically consistent opinion phrases}, and 
(3) an edge \mbox{$(n_i\rightarrow n_j) \in E$} represents a explanation relationship from $n_i$ to $n_j$. That is, the member phrases of $n_i$ explain the member phrases of $n_j$.\smallskip

The right of Figure~\ref{fig:example} depicts an opinion graph obtained from the opinion phrases mined from the given reviews. Observe that a ``perfect node'' would contain paraphrases and two perfect nodes $n_i$ and $n_j$ will be connected with an edge \mbox{$(n_i\rightarrow n_j) \in E$} if and only if $(p_i\rightarrow p_j) \; \forall \; p_i \in n_i, \; p_j \in n_j$, as is the case in the example of Figure 1. In practice, however, we often deal with imperfect nodes, containing semantically \textsl{similar} phrases and we draw an edge between two nodes whenever
an explanation relationship between the two nodes is very likely (i.e., when a significant number of explanation relationships exist between opinions in the two nodes).

Our goal is to build an opinion graph $G=(N,E)$ with optimal precision and recall for both the nodes (i.e., the clusters of opinion phrases) and the edges (i.e., the explanations between clusters). It is hard, if at all possible, to produce an opinion graph with a single end-to-end model because of the need for mining both opinion phrases and their relationships, as well as the scale of the problem, which often involves thousands of reviews. Therefore, just like the knowledge base construction pipelines~\cite{fader2011identifying, weikum2010information, dong2014knowledge}, we decompose the opinion graph construction problem into several sub-problems and focus on optimizing each sub-problem individually.

\subsection{Opinion Graph Construction Pipeline}
Our opinion graph construction method is inspired by methods used in 
knowledge base construction. We break down the construction of an opinion graph into the four components as illustrated in Figure \ref{fig:overview}. We provide an overview of each component next.\smallskip

\noindent
\textbf{Opinion Mining~} The first step mines opinion phrases from a set of reviews about an entity. For this, we can leverage Aspect-based Sentiment Analysis (ABSA) models~\cite{pontiki2015semeval,pontiki2016semeval} and, in our pipeline, we use an open-source system~\cite{Li:2019:Opine}. The system also predicts the aspect category and sentiment associated with every opinion phrase. As we describe in Section~\ref{sec:cluster}, we exploit these additional signals to improve opinion phrase canonicalization. \smallskip

\noindent    
\textbf{Explanation Mining~}
Next, \system{} discovers explanation relationships, if any, between pairs of extracted opinion phrases from reviews.
We use crowdsourcing to obtain domain-specific labeled data, and develop a supervised multi-task classifier
to discover the explanation relationship between two opinion phrases. Our model outperforms a series of baseline approaches~\cite{rocktaschel2015reasoning,parikh-etal-2016-decomposable}, including the fine-tuned BERT model~\cite{devlin2018bert}. 
    
\noindent \textbf{Opinion Phrase Canonicalization~}  Semantically similar opinion phrases are grouped together (e.g., ``\asop{not far away from Fisherman's Wharf}'' and ``\asop{close to the wharf}'') to form a node in the opinion graph. This is necessary as reviews  
overlap significantly in content and, hence, contain many similar opinion phrases.
To canonicalize opinion phrases, we develop a novel opinion phrase representation learning framework that learns opinion phrase embeddings using weak supervision obtained from the previous steps, namely predicted aspect categories, sentiment polarity scores, and explanation relationships. Similar to entity canonicalization techniques for open knowledge base construction~\cite{Vahishth:2018:CESI,Chen:2019:CanonicalizingKB}, we apply a clustering algorithm to the learned opinion phrase embeddings to cluster semantically similar opinion phrases to canonicalize those opinion phrases. 
We demonstrate improvements in the quality of the canonicalization outcome using our learned opinion phrase embeddings.

\smallskip

\noindent \textbf{Opinion Graph Generation~} Finally, we present an algorithm to construct the final opinion graph. The algorithm constructs an opinion graph by connecting graph nodes according to the aggregated explanation predictions between opinion phrases in the respective nodes. Our user study shows that our method produces graphs that are both accurate and intuitive.

\begin{figure*}[t]%
    \centering
    \subfloat[Phase-one annotation ask]{{\includegraphics[width=.47\linewidth,valign=t]{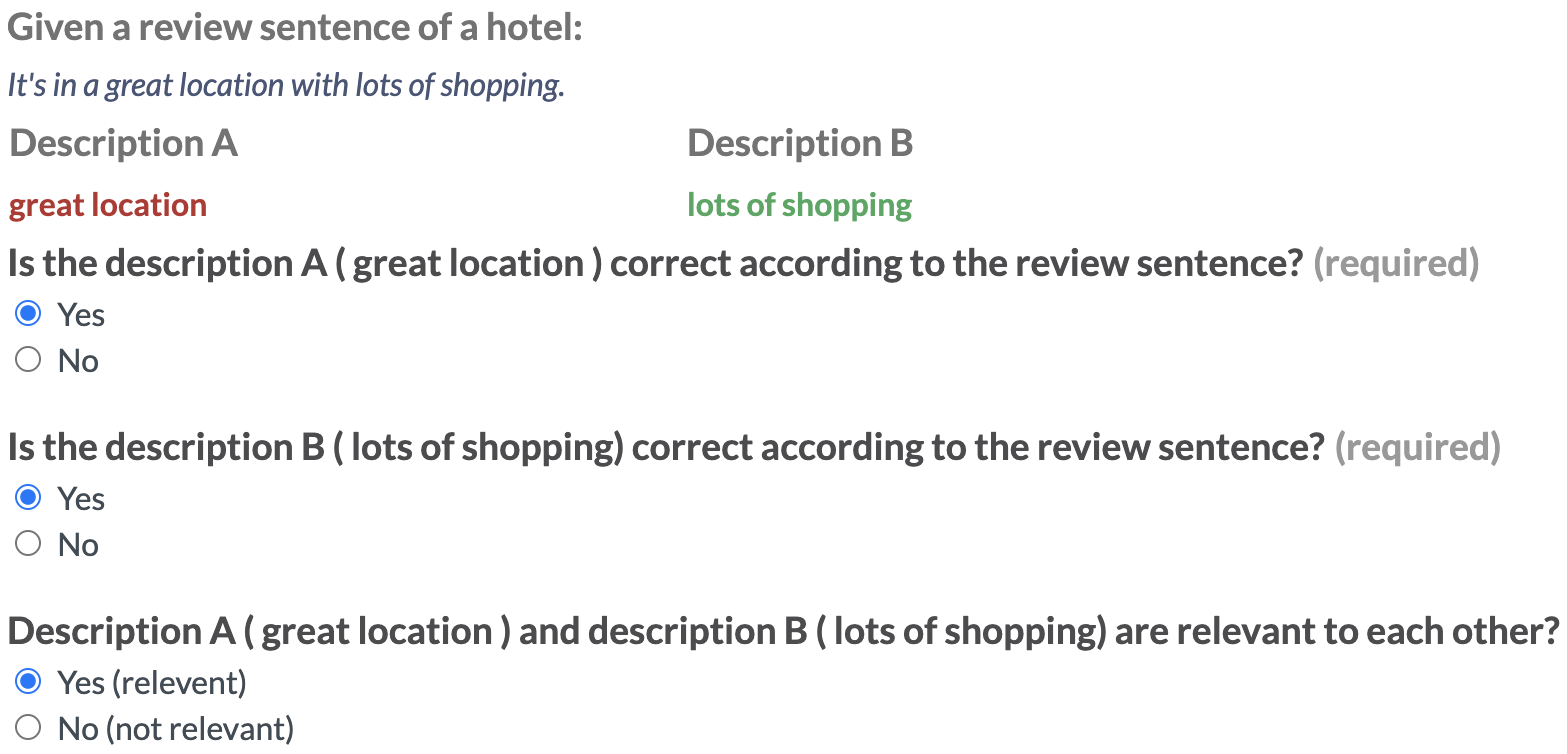} }}\hfill
    \subfloat[Phase-two annotation task]{{\includegraphics[width=.43\linewidth,valign=t]{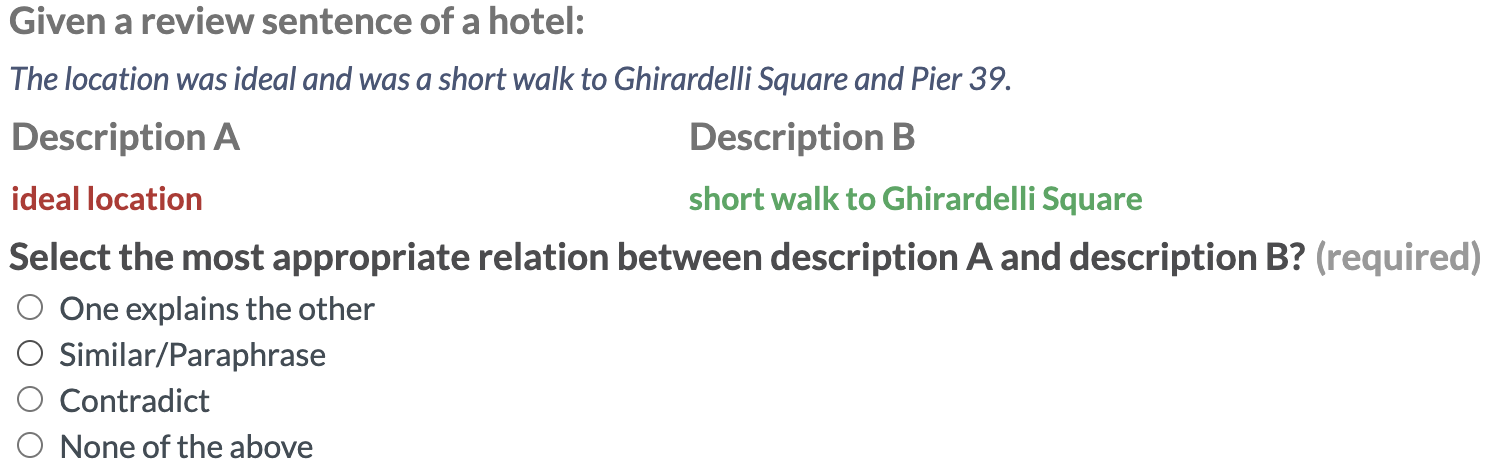} }\vphantom{\includegraphics[width=.49\linewidth,valign=t]{figs/appen_task1.png}}}%
    \vspace{-2mm}
    \caption{Human annotation tasks for explanation mining.}%
    \label{fig:appen:train}%
\vspace{-2mm}
\end{figure*}

\section{Mining Explanations}\label{sec:exp}
A significant task underlying the construction of an opinion graph is to determine when one opinion phrase explains another. 
For example, \textsl{``close to Muni bus stops''} is an explanation of \textsl{``convenient location''}, but is not an explanation for \textsl{``close to local attractions''}. 
Similarly, \textsl{``on a busy main thoroughfare''} is an explanation for \textsl{``very noisy rooms''} but not necessarily an explanation for \textsl{``convenient location''}. 

Mining explanations between opinion phrases from reviews is related to two problems: entity relation classification~\cite{zhou2016attention, wu2019enriching} (or relation extraction) and recognizing textual entailment (RTE)~\cite{Dagan:2005:PascalRTE}. The entity relation classification problem takes a sequence of text and a pair of entities as the input and learns to classify the relationship between the entities with domain-specific training data. As the models are trained and tailored by domain-specific tasks, it is infeasible to directly train the entity relation classification models for our explanation mining task. Recognizing textual entailment (RTE) problem, on the other hand, considers two sequences of text, often referred as premise and hypothesis, and determines whether the hypothesis can be inferred from the premise. Although it also considers the inference relationships between two pieces of text, RTE models trained over general text are still inadequate for mining explanations from reviews. This is based on two observations: (a) domain-specific knowledge is often necessary to understand the nuances of opinion relationships; (b) in many cases, having access to the full review is crucial to judge potential explanations. 
In fact, we evaluated a state-of-the-art RTE model~\cite{parikh-etal-2016-decomposable} trained on open-domain data~\cite{snli:emnlp2015} and observed a very low explanation accuracy of $34.3\%$. In Section ~\ref{sec:eval}, we re-trained both entity relation classification model and RTE models on the review domain and confirmed that they are still less accurate than our proposed model.

In what follows, we first describe how we collect domain-specific data for training through crowd-sourcing and then present our multi-task classifier.

\subsection{Collecting Human Annotations}

We use a two-phase procedure to collect two domain-specific training datasets for the hotel and restaurant domains.
The goal of the first phase is to prune
pairs of opinion phrases that are \textit{irrelevant} to each other. In the second phase, the crowd workers label the remaining relevant pairs of opinion phrases. That is, given a pair of relevant opinion phrases, we ask crowd workers to determine if one opinion phrase explains another. If the answer is yes, we ask them to label the direction of the explanation. In both phases, 
we provide as context, the review where the opinion phrases co-occur
to assist crowd workers in understanding the opinion phrases and hence, make better judgments. Figure~\ref{fig:appen:train} demonstrates two example tasks for each phase respectively. 

We obtained our training data via the Appen crowdsourcing platform\footnote{\url{https://appen.com/}}. To control the quality of the labels, we selected crowdworkers who can achieve at least $70\%$ accuracy on our test questions. For each question, we acquire 3 judgments and determine the final label via majority vote. For the first phase, we hired $832$ crowdworkers and observed a $0.4036$ Fleiss' kappa inter-annotator agreement rate~\cite{fleiss2013statistical}; for the second phase, we hired $322$ crowdworkers with a $0.4037$ inter-annotator agreement rate. 
As opposed to obtaining labels through a single phase, our two-phase procedure breaks down the amount of work for each label into smaller tasks and therefore renders higher quality annotated data. This is confirmed by our trial run of a single-phase procedure, which only recorded a $0.0800$ inter-annotator agreement rate.

We obtained $19K$ labeled examples this way with $20\%$ positive examples (i.e., opinion phrases in an explanation relationship). For our experiment, we used a balanced dataset with $7.4K$ examples.

\subsection{Explanation Classifier}\label{sec:exp:data}
We observe that the context surrounding the opinion phrases and the word-by-word alignments between the opinion phrases are very useful for our explanation mining task. 
For example, the opinion phrases \asop{``noisy room''} and \asop{``right above Kearny St''}, may appear to be irrelevant to each other since one is about \textsl{room quietness} and the other is about \textsl{location}. However,  the context in the review where they co-occur, \asop{``Our room was noisy. It is right above Kearny St.''} allows us to conclude that \asop{``right above Kearny St''} is an explanation for \asop{``noisy room''}. In addition to context, the word-by-word alignments between opinion phrases can also be very beneficial for explanation mining. For example, from two phrases, \asop{``easy access to public transportation''} and \asop{``convenient location''}, the word-by-word alignments between \asop{``easy access to''} and \asop{``convenient''}, as well as \asop{``public transportation''} and \asop{``location''} makes it much easier to determine that the first phrase forms an explanation to the latter one. 

However, existing models do not incorporate both types of information. Relation extraction models for constructing knowledge bases primarily focus on capturing the context between the given opinion phrases and they rarely explicitly consider word-by-word alignments between opinion phrases. RTE models mainly concentrate on aligning words between two pieces of text in the input and ignores the context.

\begin{figure}[!t]
    \centering
    \includegraphics[width=0.47\textwidth]{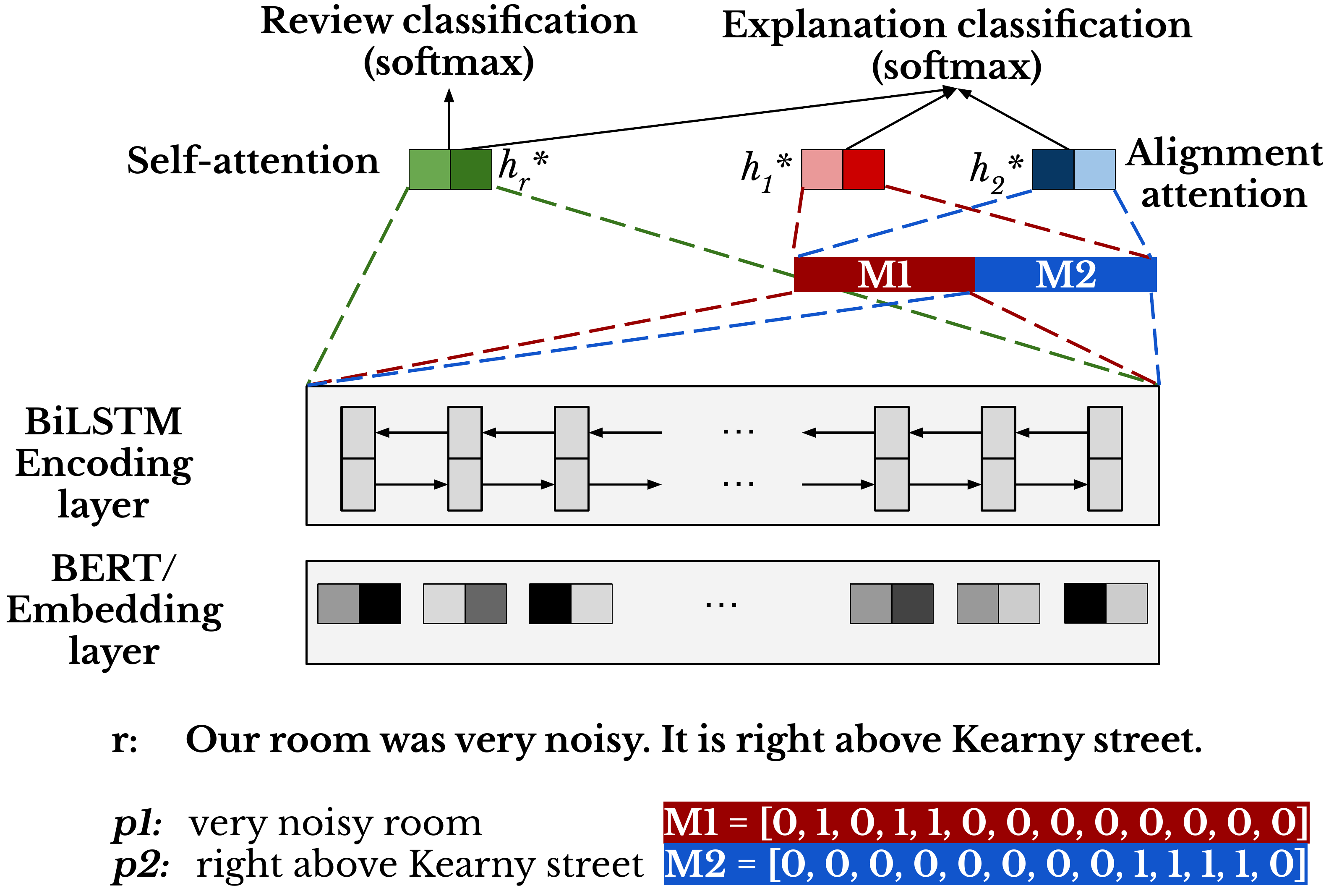}
    \caption{Model architecture of \expcls.}
    \label{fig:expmodel}
\vspace{-3mm}
\end{figure}

Therefore, to mine the explanations more effectively, we design a multi-task learning model, which we call MaskedDualAttn, for two classification tasks: (1) {\em Review classification}: whether the review contains explanations; (2) {\em Explanation classification}: whether the first opinion phrase explains the second one  (Figure~\ref{fig:expmodel}). Intuitively, we want the model to capture signals from the context and the opinion phrases. 
Our technique, which accounts for both the context surrounding the opinion phrases and the word-by-word alignments between opinion phrases, is a departure from prior methods in open-domain RTE and entity relation classification, which do not consider both information at the same time. Our ablation study confirms that both tasks are essential for mining explanations effectively (Section~\ref{subsec:eval_explanation}).
Table \ref{tab:symbol1} summarizes the notations used in this section.

\smallskip
\noindent \textbf{Input and Phrase Masks.} 
The input to the classifier consists of a review $r = (w_1, ..., w_L)$ with $L$ words, and two opinion phrases, $p_i$ and $p_j$. 
For each phrase $p=(o, a)$, we create a binary mask, \mbox{$\mathbf{m} = (m_1, ..., m_L)$}, 
which allows the model to selectively use the relevant parts of the full review encoding:

\vspace{-2mm}
\begin{equation*}
m_i = 
\begin{cases}
    1, ~~\text{if~} w_i \in a\cup o \\
    0, ~~\text{otherwise}.
\end{cases}
\end{equation*}
\vspace{-2mm}

\noindent We denote the binary masks for $p_i$ and $p_j$ as $\mathbf{m}_i$ and $\mathbf{m}_j$ respectively.

\smallskip
\noindent \textbf{Encoding.} We first encode tokens in the review $r$ through  
an embedding layer, followed by a BiLSTM layer. We denote the output vectors from the BiLSTM layer as $H=[h_1, ..., h_L] \in \mathbb{R}^{k\times L}$, where $k$ is a hyperparameter of the hidden layer dimension. We do not encode the two opinion phrases separately, but mask the review encoding using $\mathbf{m}_i$ and $\mathbf{m}_j$. Note that we can also replace the first embedding layer with one of the pre-trained models, e.g., BERT~\cite{devlin2018bert}. Our experiment demonstrates that using BERT is able to further improve the performance by $4\%$ compared to a word2vec embedding layer.

\smallskip
\noindent \textbf{Self-attention.} 
There are common linguistic patterns for expressing explanations. A simple example is the use of connectives such as ``because'' or ``due to''.  
To capture the linguistic features used to express explanations, we use the self-attention mechanism~\cite{bahdanau2014neural}, which is a common technique to aggregate hidden representations for classification:

\vspace{-1mm}
\begin{align}
    &\mathrm{M} = \text{tanh}(\mathrm{W}^\mathrm{H} H + \mathrm{b}^\mathrm{H})  &\mathrm{M} &\in \mathbb{R}^{k\times L}\\
    &\alpha = \text{softmax}(\mathrm{w}^\mathrm{T} \mathrm{M}) &\alpha & \in \mathbb{R}^{L}\\
    &\mathrm{h}_r^* = \text{tanh} (H \mathrm{\alpha}^\mathrm{T}) & \mathrm{h}_r^*& \in \mathbb{R}^{k},
\end{align}
where $\mathrm{W}^\mathrm{H}\in \mathbb{R}^{k\times k}$,  $\mathrm{b}^\mathrm{H}, \mathrm{w} \in  \mathbb{R}^{k}$ are three trainable parameters. We obtain the final sentence representation as $\mathrm{h}_r^*$.

\smallskip
\noindent \textbf{Alignment attention. }
Although the self-attention mechanism has a general capability of handling linguistic patterns, we consider it is insufficient to accurately predict the explanation relationship between opinion phrases. Thus, we implement another {\it alignment attention} layer to directly capture the similarity between opinion phrases.
Our alignment attention only focuses on opinion phrases, which is different from the self-attention layer that considers all input tokens, and it has a two-way word-by-word attention mechanism~\cite{rocktaschel2015reasoning} to produce a soft alignment between words in the input opinion phrases. To align $p_i$ with $p_j$, for each word $w_t\in p_i$, we get a weight vector $\alpha_t$ over words in $p_j$ as follows:

\begin{align}
& d_t = \mathrm{U}^\mathrm{h} h_t + \mathrm{U}^\mathrm{r} r_{t-1} & d_t & \in \mathbb{R}^{k}\\
& \mathrm{M}_t = \text{tanh}(\mathrm{U}^\mathrm{H} H + \underbrace{[d_t;...;d_t])}_{L \text{\em\ times}}) & \mathrm{M}_t& \in  \mathbb{R}^{k\times L}\\
& \alpha_t = \text{softmax}(\mathrm{u}^\mathrm{T} \mathrm{M}_t - c \overline{\mathbf{m}_j}) &  \alpha_t &\in \mathbb{R}^{L}\\
& r_t = H \alpha_t + \text{tanh}(\mathrm{U}^\mathrm{t} r_{t-1}) &  r_t &\in \mathbb{R}^{k},
\end{align}

where $\mathrm{U}^\mathrm{H}, \mathrm{U}^\mathrm{h}, \mathrm{U}^\mathrm{r}, \mathrm{U}^\mathrm{t} \in \mathbb{R}^{k\times k}$ and $\mathrm{u} \in \mathbb{R}^{k}$ are five trainable parameters; $h_t \in \mathbb{R}^{k}$ is the $t$-th output hidden state of $H$; $r_{t-1} \in \mathbb{R}^{k}$ is the representation of the previous word; and $\overline{\textbf{m}_j}$ is the reversed binary mask tensor for $p_j$. The final presentation of opinion phrase $p_i$ is obtained from a non-linear combination of $p_i$'s last hidden state $h_{|p_i|}$ and last output vector $r_{|p_i|}$:
\begin{equation}
    \mathrm{h}_i^* = \text{tanh}(\mathrm{U}^{\mathrm{x}} r_{|p_i|} + \mathrm{U}^{\mathrm{y}} h_{|p_i|}) \qquad\qquad\quad \mathrm{h}_i^* \in \mathbb{R}^{k},
\end{equation}
where $\mathrm{U}^{\mathrm{x}}, \mathrm{U}^{\mathrm{y}} \in \mathbb{R}^{k\times k}$ are two trainable parameters. 
Similar to the above mentioned procedure for aligning opinion phrase $p_i$ from $p_j$, we also align opinion phrase $p_j$ from $p_i$ and obtain its final representation $\mathrm{h}_j^*$.

\setlength{\tabcolsep}{2pt}
\begin{table}[!t]
    \small
    \centering
    \begin{tabular}{cc}
    \toprule
        \textbf{Notation} & \textbf{Meaning} \\ \midrule
        $p = (o, a)$ & Opinion phrase, opinion term, and aspect term \\
        $L$ & Sequence length of a review $r$ \\
        $\mathbf{m}_i$, $\mathbf{m}_j$ & Binary masks of the phrases $p_i$ and $p_j$ \\ 
        $k$ & Size of hidden states \\
        $H=[h_1, ..., h_L]$ & Output hidden states from BiLSTM layer \\
        $\mathrm{W}^\mathrm{H}, \mathrm{b}^\mathrm{H}, \mathrm{w}$ & Self-attention trainable weights \\
        $\mathrm{U}^\mathrm{H}, \mathrm{U}^\mathrm{h}, \mathrm{U}^\mathrm{r}, \mathrm{U}^\mathrm{t}, \mathrm{u}, \mathrm{U}^{\mathrm{x}}, \mathrm{U}^{\mathrm{y}}$ & Alignment-attention trainable weights \\
        \bottomrule
    \end{tabular}
    \caption{Notations for explanation classifier.}
    \label{tab:symbol1}
    \vspace{-3mm}
\end{table}

\smallskip
\noindent \textbf{Prediction and Training.} 
The probability distributions for the review classification ($\mathbf{s}_r$) and explanation classification 
($\mathbf{s}_e$) tasks are obtained from two softmax classifiers respectively:
\begin{align}
    &\mathbf{s}_r = \text{softmax}(\mathrm{W}^{\mathrm{r}} \mathrm{h}_r^* + \mathrm{b}^{\mathrm{r}}) & \mathbf{s}_r& \in \mathbb{R}^{2}\\
    &\mathbf{s}_e = \text{softmax}(\mathrm{W}^{\mathrm{e}} \mathrm{h}_e^* + \mathrm{b}^{\mathrm{e}}) & \mathbf{s}_e& \in \mathbb{R}^{2}, \label{eq:e}
\end{align}
where $\mathrm{h}_e^*= [\mathrm{h}_r^* ; \mathrm{h}_i^* ; \mathrm{h}_j^* ]$ is the concatenation of the sentence's and opinion phrases' representations;
$\mathrm{W}^{\mathrm{r}} \in \mathbb{R}^{2\times k}, \mathrm{b}^{\mathrm{r}} \in \mathbb{R}^2$ and 
$\mathrm{W}^{\mathrm{e}} \in \mathbb{R}^{2\times k}, \mathrm{b}^{\mathrm{e}} \in \mathbb{R}^2$ are the classifiers' weights and biases respectively.
Finally, we define the training objective $J$ as follows:
\vspace{-1mm}
\begin{equation}
J = J_o + \lambda J_r,
\end{equation}
where $J_r$ and $J_o$ are the cross-entropy loss for the first and second classification task, respectively;
$\lambda$ is a tunable hyper-parameter.
\section{Canonicalizing Opinion Phrases}\label{sec:cluster}

The goal of this component is to group duplicates or very similar opinion phrases together in order to build a concise opinion graph. We call this process {\em canonicalizing opinion phrases}. This is a necessary step as reviews contain 
a variety of linguistic variations to express the same or similar opinions. 
For example, \asop{``one block from beach''}, \asop{``close to the pacific ocean''}, \asop{``unbeatable beach access''}, and \asop{``very close to the sea''} are different phrases used in hotel reviews to describe the same opinion. Other examples (e.g., \asop{``great location''} and \asop{``good location''}) are shown on the right of Figure~\ref{fig:example}.

A widely used method for representing a phrase is the average word embeddings of the phrase based on a pre-trained word embedding model (e.g., GloVe)~\cite{kusner2015word, Vahishth:2018:CESI}.
However, a serious limitation of this approach is that it may wrongly cluster opinion phrases that share the same opinion/aspect term but are not necessarily similar. For example, the average word embeddings for the opinion phrase ``\asop{very close to the ocean}'' is closer to an irrelevant opinion phrase ``\asop{very close to the trams}'' than a semantically similar opinion phrase ``\asop{2 mins walk to the beach}''. See Figure~\ref{fig:emb}(a) for more examples.

To account for semantic similarity, we consider opinion phrase representation learning to learn opinion phrase embeddings before applying
a clustering algorithm such as $k$-means to group similar opinion phrases together. Our method offers 
two major benefits. First, it utilizes only weak supervision: our model does not require any additional labels for training as it leverages the outputs of previous components, namely the aspect categories and sentiment polarity of opinion phrases (in Opinion Mining), and mined explanations (in Explanation Mining). Second, our method allows us to use existing clustering algorithms, which we improve by simply using our opinion phrase representations as features.

\subsection{Opinion Phrase Representation Learning}

We develop an opinion phrase representation learning framework {\bf W}eakly-{\bf S}upervised {\bf O}pinion {\bf P}hrase {\bf E}mbeddings (\canonical), which has two key properties: (1) different embeddings are used for opinion and aspect terms separately, which are then merged into an opinion phrase embedding, and (2) it uses weak supervision to incorporate the semantic meaning of opinion phrases into the opinion phrase embeddings by minimizing the vector reconstruction loss, as well as additional losses based on predicted aspect category, sentiment polarity, and explanations obtained from the previous steps of \system.
The first idea makes it easier for the model to learn to distinguish lexically similar but semantically different opinion phrases. For example, the model can distinguish \asop{``very close to the trams''} and \asop{``very close to the ocean''} based on the aspects \asop{``tram''} and \asop{``ocean''}, which have different representations. The second idea enables the learning of opinion phrase embeddings without additional cost. With the additional loss functions based on signals extracted in the previous steps, the model can incorporate sentiment information into opinion phrase embeddings while retaining the explanation relationship between opinion phrases in the embedding space.

\begin{figure*}[t]
\begin{minipage}{0.99\textwidth}
\centering
\noindent
\includegraphics[width=0.9\linewidth]{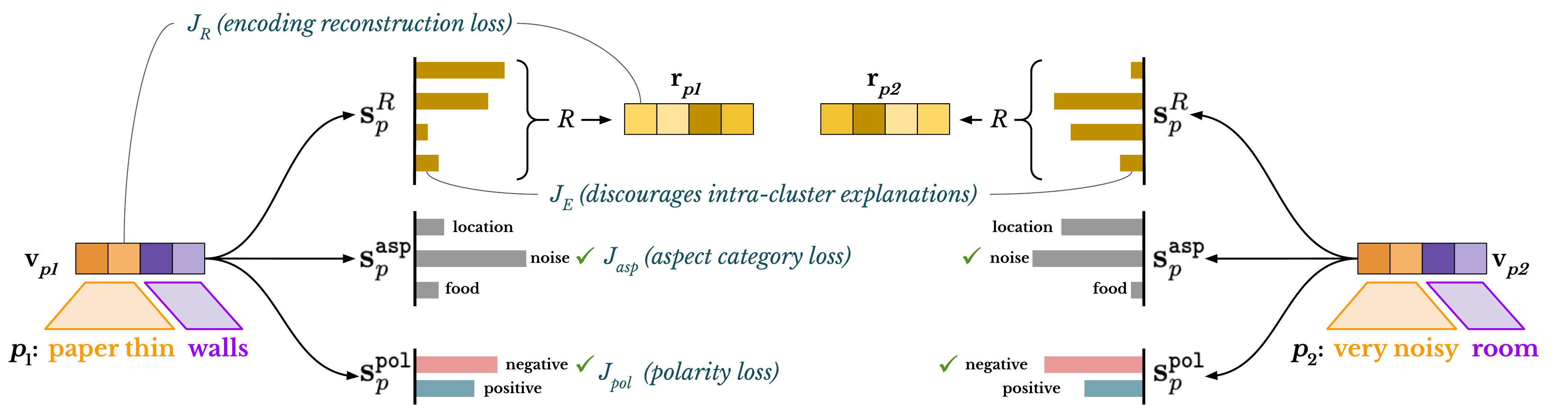}
\caption{Overview and loss functions of our opinion phrase representation learning framework (\canonical).}
\label{fig:clustermodel}
\end{minipage}
\vspace{-3mm}
\end{figure*}

Figure~\ref{fig:clustermodel} illustrates 
the learning framework of \canonical. The model encodes an opinion phrase into an embedding vector, which is the concatenation of an opinion term embedding and an aspect term embedding. Then, the opinion phrase embedding is used as input to evaluate multiple loss functions. The total loss is used to update the model parameters, including the opinion phrase embeddings themselves (i.e., opinion and aspect embeddings). We describe each component and loss function next. Table \ref{tab:symbol2} summarizes the notations, which we will use in this section.

\setlength{\tabcolsep}{2pt}
\begin{table}[!t]
    \small
    \centering
    \begin{tabular}{cc}
    \toprule
        \textbf{Notation} & \textbf{Meaning} \\ \midrule
        $P=\{p_i\}_{i=1}^{N_p}$ & Input opinion phrases \\
        $E=\{e_i\}_{i=1}^{N_e}$ & Explanations extracted by Section~\ref{sec:exp} \\
        $\mathrm{v}_{p} = [\mathrm{v}_a; \mathrm{v}_o]$ & Embeddings for opinion phrase, opinion term, aspect term\\
        $\mathrm{W}^{\mathrm{a}}$ & Trainable parameter for opinion phrase encoding\\
        $\mathrm{W}^{\mathrm{R}}, \mathrm{b}^{\mathrm{R}}$ & Trainable parameters for opinion phrase reconstruction\\
        $J_R $ & Reconstruction loss \\
       $\mathrm{W}^\mathtt{asp}, \mathrm{b}^\mathtt{asp} $ & Trainable parameters for aspect classification \\
       $\mathrm{W}^\mathtt{pol}, \mathrm{b}^\mathtt{pol}$ & Trainable parameters for polarity classification\\
        $J_{asp}, J_{pol}$ & Aspect and polarity classification loss\\ 
        $J_E$ & Intra-cluster explanation loss \\
        \bottomrule
    \end{tabular}
    \caption{Notations for opinion representation learning.}
    \label{tab:symbol2}
    \vspace{-4mm}
\end{table}

\smallskip
\noindent \textbf{Input.} The input to the model is a set of $N_p$ opinion phrases $P = \{p_i\}_{i=1}^{N_p}$ 
that are extracted from reviews about a single entity (e.g., a hotel), and a set of $N_e$ explanations $E_e = \{e_i\}_{i=1}^{N_e}$ for the opinion phrases.
Recall that each opinion phrase\footnote{We omit the index $i$ of the opinion phrase and explanation below since it is clear from the context.} $p$ consists of two sequences of tokens $(o, a)$ of the opinion term and the aspect term.
We use $\mathtt{asp}$ and $\mathtt{pol}$ to denote the aspect category and sentiment labels of a phrase, predicted by  
the ABSA model during the opinion mining stage.

\smallskip
\noindent \textbf{Opinion Phrase Encoding.} Given an opinion phrase $(o, a)$, 
we first use an embedding layer with the self-attention mechanism~\cite{he2017unsupervised} to compute an aspect embedding $\mathrm{v}_a$ and an opinion embedding $\mathrm{v}_o$ respectively\footnote{The embedding architecture is similar to that of Attention-Based Auto-Encoder (ABAE)~\cite{he2017unsupervised} but our model has different encoders for aspect and opinion terms, whereas ABAE does not distinguish aspect and opinion terms and directly encodes an opinion phrase into an embedding vector.}.
The aspect embedding $\mathrm{v}_a$ is obtained by attending over the aspect term tokens $a = (w_1, \dots, w_n)$:
\begin{align}
    &u_i = \mathrm{v}_{w_i}^T \mathrm{W}^{\mathrm{a}}  \mathrm{v}_{a}' & u_i &\in \mathbb{R} \\
    &c_{i} = \frac{\exp(u_i)}{\sum_{j=1}^{m} \exp(u_j)} & c_i& \in \mathbb{R} \\ 
    &\mathrm{v}_a = \sum_{i=1}^{m} c_i \mathrm{v}_{w_i} & \mathrm{v}_a&\in \mathbb{R}^{d},
\end{align}

\noindent
where $\mathrm{v}_{w_i} \in \mathbb{R}^d$ is the output of the embedding layer for word $w_i$, $\mathrm{v}_{a}' \in \mathbb{R}^d$ is the average word embedding for words in $a$, and $\mathrm{W}^{\mathrm{a}} \in \mathbb{R}^{d\times d}$ is a trainable parameter used to calculate attention weights.
We encode the opinion term $o$ into $\mathrm{v}_o$ in the same manner.
Then, we concatenate the two embedding vectors into a single opinion phrase embedding $\mathrm{v}_{p}$:
\begin{equation}
    \mathrm{v}_{p} = [\mathrm{v}_a; \mathrm{v}_o] \qquad\qquad\qquad\qquad \mathrm{v}_{p}\in \mathbb{R}^{2d}.
\end{equation}

\smallskip
\noindent \textbf{Reconstruction loss.} As in the standard auto-encoder paradigm, the main idea behind the reconstruction loss is to learn input vectors $\mathrm{v}_p$ so that they can be easily reconstructed from a representative matrix $\mathrm{R}$. Following previous studies~\cite{he2017unsupervised,angelidis2018summarizing}, we set the $K$ rows of $\mathrm{R} \in \mathbb{R}^{K\times 2d}$ using a clustering algorithm (e.g., $k$-means) over initial opinion phrase embeddings, such that every row corresponds to a cluster centroid\footnote{We freeze $\mathrm{R}$ during training after initialization to facilitate training stability, as suggested in \cite{angelidis2018summarizing}.}.
To reconstruct the phrase vector $\mathrm{v}_{p}$, we first feed it to a softmax classifier to obtain a  probability distribution over the $K$ rows of $\mathrm{R}$:
\begin{equation}\label{eq:assignment_dist}
    \mathbf{s}^R_p = \text{softmax}(\mathrm{W}^{\mathrm{R}}  \mathrm{v}_{p} + \mathrm{b}^{\mathrm{R}})\qquad\qquad \mathbf{s}^R_p \in \mathbb{R}^{K},
\end{equation}

where $\mathrm{W}^{\mathrm{R}} \in \mathbb{R}^{K\times 2d}, \mathrm{b}^{\mathrm{R}}\in \mathbb{R}^K$ are the weight and bias parameters of the classifier respectively.
We get the \textsl{reconstructed} vector $\mathrm{r}_{p}$ for opinion phrase $p$ as follows:
\begin{equation}
    \mathrm{r}_{p} = \mathrm{R}^T \mathbf{s}^R_p \qquad\qquad \qquad\qquad \qquad  \mathrm{r}_{p}\in \mathbb{R}^{2d}.
\end{equation}

We use the triplet margin loss~\cite{balntas2016learning}
as the cost function, which moves the input opinion phrase $\mathrm{v}_p$ closer to the reconstruction $\mathrm{r}_{p}$, and further away from $k_n$ randomly sampled negative examples:
\begin{equation}
    J_R = \sum_{p \in P} \sum_{i=1}^{k_n}\text{max}(0, 1- \mathrm{r}_{p} \mathrm{v}_{p}+\mathrm{r}_{p} \mathrm{v}_{n_i}),
\end{equation}
where $n_i \in P$ are randomly selected negative examples.
The sampling procedure tries to sample opinion phrases that are not similar to the input opinion phrase with respect to the probability distribution of Eq.~(\ref{eq:assignment_dist}).
For an opinion phrase $p$, the probability of another opinion phrase $p'$ being selected as a pseudo negative example is \textsl{inversely} proportional to the cosine similarity between $\mathbf{s}^R_p $ and $\mathbf{s}^R_{p'}$.

\smallskip
\noindent \textbf{Aspect category and polarity loss.} We also leverage additional signals that we collected from the previous steps to obtain better representations.
For example, we would like to avoid opinion phrases such as ``\asop{friendly staff}'' and ``\asop{unfriendly staff}'' from being close in the embedding space. 
Hence, we incorporate sentiment polarity and aspect category into the framework to learn better opinion phrase embeddings with respect to sentiment information.

In \canonical{}, we add two classification objectives to learn the parameters. Specifically, we feed an opinion phrase embedding $\mathrm{v}_{p}$ into two softmax classifiers to predict the probability distributions of the aspect category and the sentiment polarity respectively:
\begin{align}
    \mathbf{s}_{p}^{\mathtt{asp}}= \text{softmax}(\mathrm{W}^\mathtt{asp}  \mathrm{v}_{p} + \mathrm{b}^\mathtt{asp})\\
    \mathbf{s}_{p}^{\mathtt{pol}}= \text{softmax}(\mathrm{W}^\mathtt{pol}  \mathrm{v}_{p} + \mathrm{b}^\mathtt{pol}).
\end{align}
The distributions $\mathbf{s}_{p}^{\mathtt{asp}}$ and $\mathbf{s}_{p}^{\mathtt{pol}}$ are used to compute cross-entropy losses $J_\mathtt{asp}$ and $J_\mathtt{pol}$ against \textsl{silver-standard} aspect and sentiment labels, predicted for each extracted phrase during opinion mining.

\smallskip
\noindent \textbf{Intra-cluster explanation loss.} Our mined explanations should also provide additional signals to learn better embeddings. Essentially, if an opinion phrase $p_i$ explains an opinion phrase $p_j$, they should belong to different clusters (i.e., they should not belong to the same cluster).
To reduce intra-cluster explanations, we define the \textsl{intra-cluster explanation loss} by the Kullback-Leibler divergence (KL) between the probability distributions $\mathbf{s}^R_{p_i}  $ and $\mathbf{s}^R_{p_j}$:

\begin{equation}
J_{E} = -\sum_{e \in E} \mathrm{KL}(\mathbf{s}^R_{p_i}, \mathbf{s}^R_{p_j}), \; e=(p_i\rightarrow p_j)\,,
\end{equation}
where $E$ is a set of pairs of opinion phrases in the mined explanations,
and $\mathrm{KL}(\cdot,\cdot)$ is the KL divergence between two distributions.

When opinion phrases $p_i$ and $p_j$ are in an explanation relationship, we would like to penalize the case where $\mathrm{KL}(\mathbf{s}^R_{p_i}, \mathbf{s}^R_{p_j})$ is small (likely to be in the same cluster).
As a result, we are able to push the embeddings of $p_i$ and $p_j$ of opinion phrases that have an explanation relationship apart from each other to discourage having intra-cluster explanations.

\smallskip
\noindent \textbf{Training objective.} We define the final loss function by combining the four loss functions defined above:
\begin{equation}\label{eq:all_loss}
J_{\text{WS-OPE}} = J_R + \lambda_\mathtt{asp} J_\mathtt{asp} + \lambda_\mathtt{pol} J_\mathtt{pol} + \lambda_E J_E,
\end{equation}
\noindent
where $\lambda_\mathtt{asp}$, $\lambda_\mathtt{pol}$, and $\lambda_E$ are three hyper-parameters to control the influence of each corresponding loss. 
In practice, we prepare two types of mini-batches; one for single opinion phrases and one for explanation pairs. For each training step, we create and use these mini-batches separately: we use the single phrase mini-batch to evaluate the reconstruction, aspect category, and polarity losses; we use the explanation mini-batch to evaluate the explanation loss. At the end of every training step, we accumulate the loss values following Eq.~(\ref{eq:all_loss}) and update the model parameters.

\smallskip
\subsection{Clustering Opinion Phrases}

After the opinion phrase representation learning, we apply a clustering algorithm over the learned opinion phrase embeddings to obtain opinion clusters. Each
opinion cluster is a node (i.e., {\it canonicalized} opinion) of the final opinion graph. 
Note that our opinion canonicalization
module is not tied to any specific
clustering algorithm. 
We will show in Section~\ref{sec:eval}, our two-stage method for generating opinion clusters performs well regardless of the choice of clustering algorithms, which also demonstrates the strength of 
opinion phrase representation learning.

We could consider directly using a score distribution $\mathbf{s}^R_p$ for clustering instead of applying a clustering algorithm to the learned opinion embeddings. However, we found that $\mathbf{s}^R_p$ does not perform well compared to our approach. This is expected, as the classifier responsible for producing $\mathbf{s}^R_p$ has only been trained via the reconstruction loss, whereas the phrase embeddings have used all of the four signals, thus producing much richer representations.
We conduct further analysis on the contributions of the multiple loss functions in \ref{sec:cluster:sensitivity}.

\section{Generating Opinion Graphs} \label{sec:graph}

Based on mined explanations and canonicalized opinions from Sections~\ref{sec:exp} and \ref{sec:cluster}, the final step for generating an opinion  graph is to predict edges between nodes. In theory, when using perfectly accurate explanations and opinion clusters, generating such edges is trivial. Intuitively, when an opinion phrase explains another opinion phrase, opinion phrases that are paraphrases of the first phrase should also explain phrases that paraphrase the latter one. In other words, given a set of explanations $E$ and two groups of opinion phrases, $n_i$ and $n_j$, there should be an edge from $n_i$ to $n_j$ if there exists an edge between two opinion phrases in $n_i$ and $n_j$ respectively:

\vspace{-1mm}
\[e=(n_i\rightarrow n_j) \text{~is~true}, \text{if~} \exists \, e=(p\rightarrow p')\in E|  p\in n_i, p'\in n_j\]
\vspace{-1mm}

For example, when we know that \asop{``close to the beach''} $\rightarrow$ \asop{``good location''}, we are able to conclude (\asop{``close to the beach''}, \asop{``near the beach''}, \asop{``walking distance to the beach''})  $\rightarrow$ (\asop{``good location''}, \asop{``great location''}, \asop{``awesome location''}).

However, in practice, our obtained explanations and nodes are not perfect. Thus, we may get a lot of false positive edges based on the above criteria. To minimize the false positives, we use a simple heuristic to further prune the edges, which is based on the observation that two groups of opinions seldom explain each other at the same time. 
\vspace{-1mm}
\[e=(n_i\rightarrow n_j) \text{~is~true}, \text{if~} \sum_{e\in E_{ij}} \mathrm{p}_e - \sum_{e'\in E_{ji}} \mathrm{p}_e' > 0\, ,\]
where $E_{ij} = \{e=(p_i\rightarrow p_j) \mid p_i\in n_i, p_j\in n_j\}$ and $E_{ji} = \{e=(p_j\rightarrow p_i) \mid p_i\in n_i, p_j\in n_j\}$ are the explanations from $n_i$ to $n_j$ and $n_j$ to $n_i$ respectively; and $\mathrm{p}_e$ and $\mathrm{p}_e'$ are the explanation probabilities obtained from our explanation mining classifier. 

Deriving edges between canonicalized opinions is a difficult problem in general. There are many ways to optimize this step further, and we leave this as part of our future work.

\begin{table}
\small
\resizebox{1.0\linewidth}{!}{
\begin{tabular}{C{1.5cm}|C{5.0cm}|c}
\toprule 
\textbf{Group} & \textbf{Models} & {{\bf Acc.}} \\\midrule
& Two-way attention  & 74.78 \\
{\sc RTE} & Decomposable attention & 76.26 \\
& RTE-BERT & 79.75  \\\midrule
{\sc RelCLS} & Sent-BiLSTM & 75.41   \\
& Sent-BERT & 81.79 \\\midrule
{\sc Proposed} & \expcls-GloVe  & 82.20  \\
               & \expcls-BERT& \textbf{86.23} \\\midrule
{\sc Ablated} & \expcls-GloVe; single-task & 78.67 (3.53 $\downarrow$) \\
& \expcls-BERT; single-task & 80.57 (5.66 	$\downarrow$)\\\bottomrule
\end{tabular}}
\caption{Explanation mining accuracy of different models. Our ablated single-task model (GloVe and BERT) are trained without the review classification objective (i.e., $\lambda = 0$).}
\label{fig:causal_result}
\vspace{-3mm}
\end{table}
\normalsize

\begin{table*}[ht!]
\begin{minipage}[t][][b]{\linewidth}
\centering
\begin{tabular}{c|c|ccc|ccc|ccc}
\toprule 
\multicolumn{2}{c|}{}&\multicolumn{3}{c|}{{\bf Homogeneity (Precision)}} & \multicolumn{3}{c|}{{\bf Completeness (Recall)}}& \multicolumn{3}{c}{{\bf V-measure (F1)}} \\\cmidrule{3-11}
\multicolumn{2}{c|}{}& {\bf k-means} & {\bf GMM} & {\bf Cor. Cluster.} & {\bf k-means} & {\bf GMM} & {\bf Cor. Cluster.} & {\bf k-means} & {\bf GMM} & {\bf Cor. Cluster.} \\\midrule
\multirow{3}{*}{\hotel} & AvgWE & 0.6695&0.6785&0.7240&0.7577&0.7728&0.6756&0.7102&0.7219&0.6985 \\
& ABAE & 0.6626&0.6628&0.6964&0.7609&0.7522&0.7113&0.7075&0.7039&0.6966  \\
& \canonical{} (ours) & {\bf 0.7073}&{\bf 0.7177}&{\bf 0.7460}&{\bf 0.8115}&{\bf 0.8184}&{\bf 0.8370}&{\bf 0.7551}&{\bf 0.7641}&{\bf 0.7848} \\\midrule
\multirow{3}{*}{\restaurant} & AvgWE & 0.5854&0.5509&0.5851&0.8168&0.7801&0.8103&0.6778&0.6413&0.6761 \\
& ABAE &0.5563&0.5553&{\bf 0.6256}&0.7927&0.7779&0.7819&0.6492&0.6432&0.6918 \\
& \canonical{} (ours) & {\bf 0.5920}&{\bf 0.5572}&0.6158&{\bf 0.8333}&{\bf 0.8111}&{\bf 0.8155}&{\bf 0.6877}&{\bf 0.6555}&{\bf 0.6985} \\
\bottomrule
\end{tabular}
\caption{Opinion phrase canonicalization performance on \hotel\ and \restaurant\ datasets.}\label{fig:cluster:all}
\end{minipage}
\vspace{-4mm}
\end{table*}
\normalsize

\section{Evaluation}\label{sec:eval}
We evaluate \system\ with three types of experiments. We use two review datasets for evaluation: a public \yelp{} corpus of $642K$ restaurant reviews and a private \hotel{} corpus\footnote{Data was collected from multiple hotel booking websites.} of $688K$ hotel reviews. 
For the mining explanations and canonicalizing opinion phrases, we perform automatic evaluation over crowdsourced gold labels\footnote{We release the labeled datasets at {\url{https://github.com/megagonlabs/explainit}}.}. To evaluate the quality of the generated opinion graph, we conducted a user study.  

\subsection{Mining Explanations}\label{subsec:eval_explanation}
\subsubsection{Dataset and metric} Based on the data collection process we described in \ref{sec:exp:data}, we used a dataset with $7.4K$ balanced examples in \hotel\ domain. We further split the labeled data into training, validation, and test sets with ratios of $(0.8, 0.1, 0.1)$. We evaluate the models by their prediction accuracy.

\subsubsection{Methods.}
We compare our explanation classifier model and baseline methods, which we categorized into three groups.
The first group ({\sc RTE}) consists of three different models for RTE, the second group ({\sc RelCLS}) consists of two models for relation classification, the third group ({\sc Proposed}) consists of different configurations of our model, and the last group ({\sc Ablated}) is for ablation study. We trained all the models on the same training data with Adam optimizer~\cite{kingma2014adam} (learning-rate=$1e-3$, $\beta_1=0.9$, $\beta_2=0.999$, and decay factor of $0.01$) for $30$ epochs. All models except BERT used the same word embedding model ({\tt glove.6B.300d})~\cite{pennington-etal-2014-glove} for the embedding layers.
For the {\sc RTE} group, the input to the models is a pair of opinion phrases. The review context information associated with the pairs is ignored.

\noindent {\bf Two-way attention}: The two-way attention model~\cite{rocktaschel2015reasoning} is a BiLSTM model with a two-way word-by-word attention, which is used in our proposed model. This can be considered a degraded version of our proposed model only with the alignment attention, which takes opinion phrases without context information.

\noindent {\bf Decomposable attention:} The decomposable attention model ~\cite{parikh-etal-2016-decomposable} is a widely used and the best non-pre-trained model for RTE tasks.

\noindent {\bf RTE-BERT}: BERT~\cite{devlin2018bert} is a pre-trained self-attention model, which is known to achieve state-of-the-art performance in many NLP tasks. We fine-tuned the BERT\textsubscript{base} model with our training data.

\smallskip

For the {\sc RelCLS} group, we follow existing relation classification techniques~\cite{zhou2016attention, lin2016neural, wu2019enriching} and formulate the explanation classification problem as a single sentence classification task. We ``highlight'' opinion phrases in a review with special position indicators~\cite{zhou2016attention} {\tt [OP1]} and {\tt [OP2]}. For example, 
``{\tt [OP1]} \asop{Good location} {\tt [OP1]} \asop{with} {\tt [OP2]} \asop{easy access to beach} {\tt [OP2]}'' highlights two opinion phrases: ``\asop{good location}'' and ``\asop{easy access to beach}''. With this input format, we can train a sentence classification model that takes into account context information while it recognizes which are opinion phrases. 

\noindent {\bf Sent-BiLSTM:} We trained a BiLSTM model with self-attention~\cite{Lin:2017:SelfAttentionLSTM}, which was originally developed for sentence classification tasks. The model architecture can be considered a degraded version of our model without the two-way word-by-word attention. Because we use special position indicators, this model classifies whether the opinion phrases are in the explanation relationship or not. 

\noindent {\bf Sent-BERT:} We fine-tuned the BERT\textsubscript{base} model for the sentence classification task with the training data. Different from RTE-BERT, Sent-BERT takes an entire review, enriched by opinion phrase markers, as the input so it can take context information into account. 

\smallskip

The last group ({\sc Proposed}) include two variations of our model:

\noindent {\bf \expcls-GloVe}: The default model with an embedding layer initialized with the GloVe ({\tt glove.6B.300d}) model.

\noindent {\bf \expcls-BERT}: We replace the embedding layer with the BERT\textsubscript{base} model to obtain contextualized word embeddings. 

\subsubsection{Result analysis}
As shown in Table~\ref{fig:causal_result},
our proposed model achieves significant improvement over baseline approaches: we largely outperform non-pre-trained textual entailment models and sentence classification models by $5.94\%$ to $7.42\%$. Furthermore, to mine explanations, models that consider context information tend to perform better. We found that BERT over sentences is $2\%$ more accurate than BERT over opinion phrases only. Lastly, leveraging pre-trained model can further improve the performance: by replacing the embedding layer with BERT, the accuracy is further improved by $4\%$.

We also conducted an ablation analysis to verify our multi-task learning framework. We tested variants of \expcls-GloVe and \expcls-BERT that were trained without the review classification objective (i.e., $\lambda=0$). The other configurations were the same as \expcls-GloVe and \expcls-BERT. The results are shown in Table~\ref{fig:causal_result} ({\sc Ablated}). From the results, we confirm that the multi-task learning significantly contributes to the performance of both of the \expcls\ models.

Since our model has both of the alignment attention and self-attention, only with the single objective function (i.e., explanation classification), the model may not be optimized well. In fact, by turning off the multi-task learning, we observe lower performance by \expcls-BERT than Sent-BERT, while \expcls-GloVe shows better performance than the BiLSTM-based baseline models (Sent-BiLSTM). 
Therefore, we consider the issue can be resolved by incorporating multiple objectives as our final models, regardless of the choice of the base model (i.e., GloVe, BERT) achieves the best performance in the explanation mining task.

\begin{figure}[t!]
    \centering
    \includegraphics[width=0.32\textwidth]{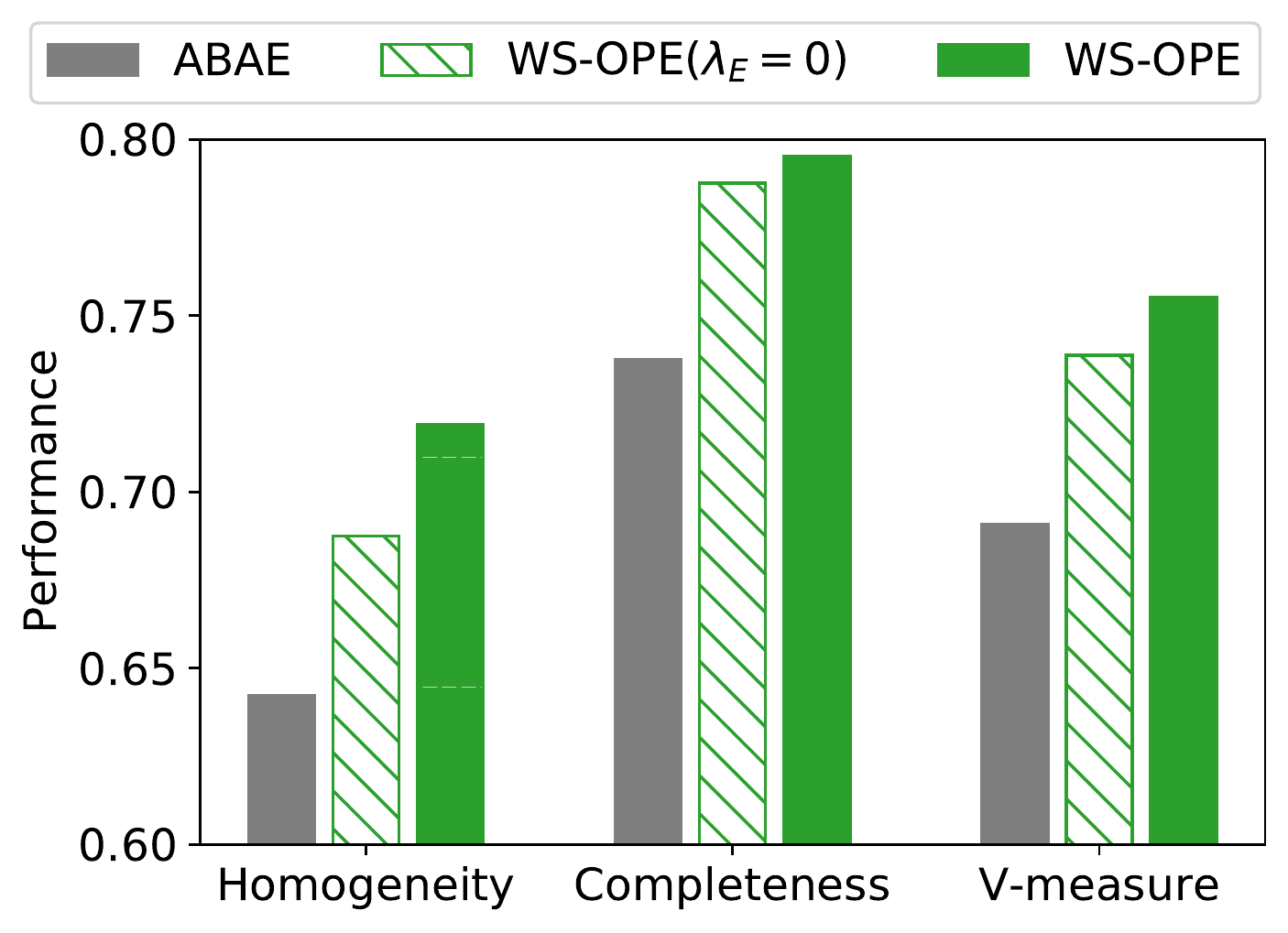}
    \caption{Ablation study on the usefulness of the intra-cluster explanation loss. Excluding the intra-cluster explanation loss (i.e., $\lambda_E=0$) hurts the opinion phrase canonicalization performance, while it still performs better than ABAE.}
    \label{fig:merge:ablation}
    \vspace{-2mm}
\end{figure}

\subsection{Canonicalizing Opinions Phrases}
\newcommand{\glove}{{AvgWE}}
\subsubsection{Dataset and metrics} For both \hotel\ and \restaurant\ domains, we first exclude entities with too few/many reviews and randomly select $10$ entities from the remaining ones. 
We also develop a non-trivial process to collect the gold clusters using crowdsourcing.  

We evaluate the performance with three metrics: (1) homogeneity, (2) completeness, and (3) V-measure, in the same manner as precision, recall, and F1-score. Homogeneity measures the {\it precision} of each cluster and scores 1.0 if each cluster contains only members of a single class. Completeness measures the {\it recall} of each true class and scores 1.0 if all members of a given class are assigned to the same cluster.
The V-measure is the harmonic mean between homogeneity and completeness scores. 

\begin{figure*}[th!]
\minipage{1.0\textwidth}
\minipage{0.3\textwidth}
  \includegraphics[width=\linewidth]{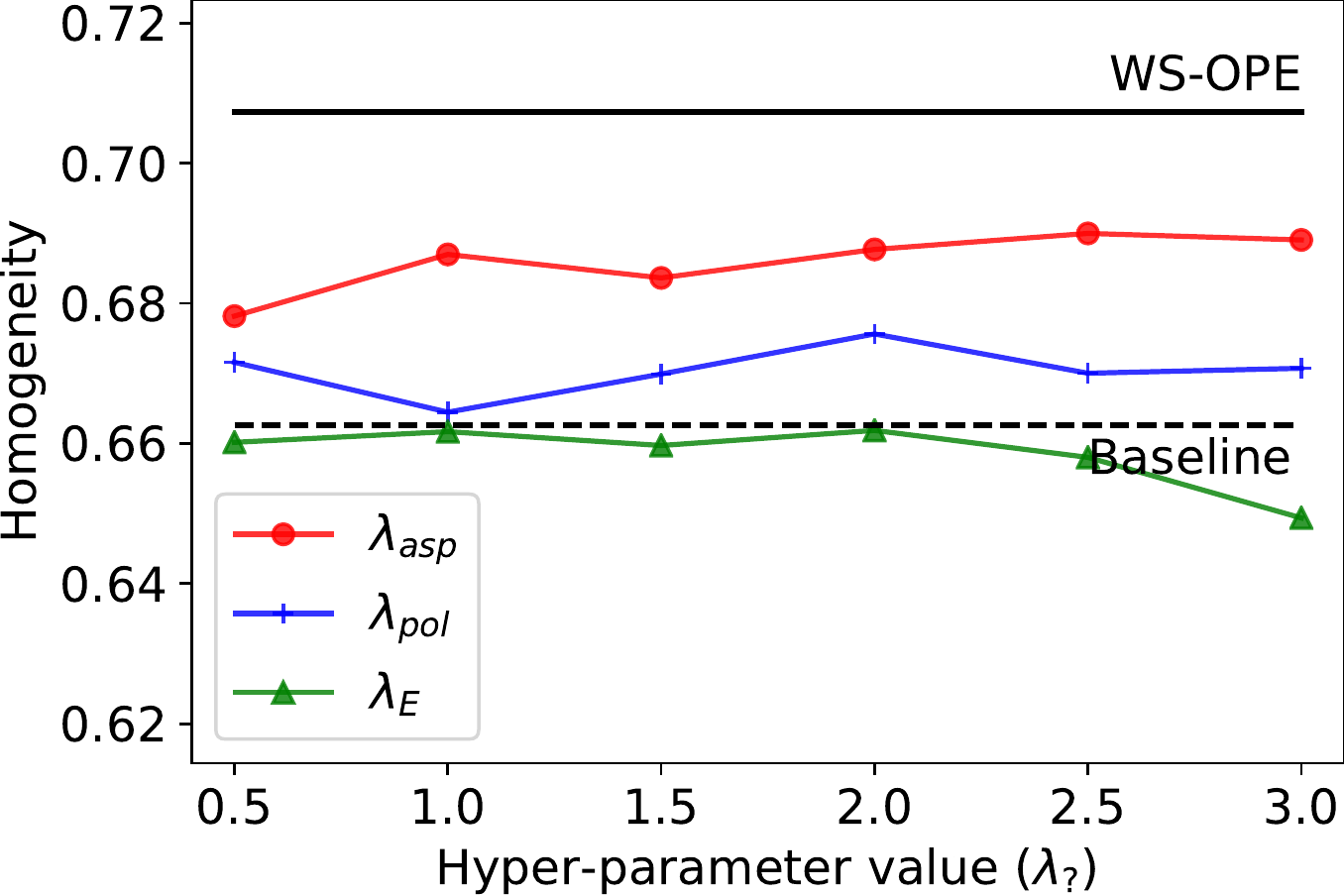}
\endminipage\hfill
\minipage{0.3\textwidth}
  \includegraphics[width=\linewidth]{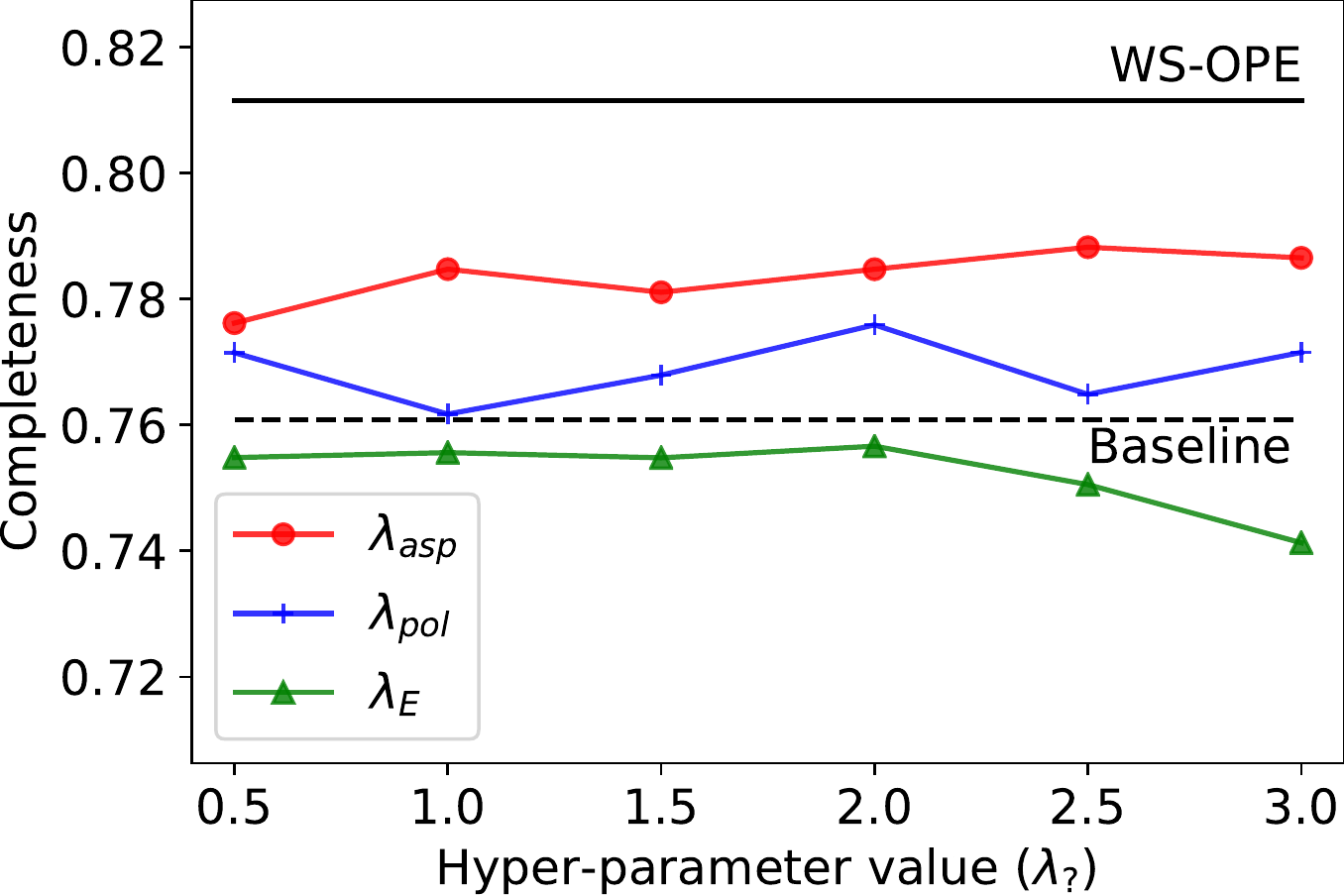}
\endminipage\hfill
\minipage{0.3\textwidth}%
  \includegraphics[width=\linewidth]{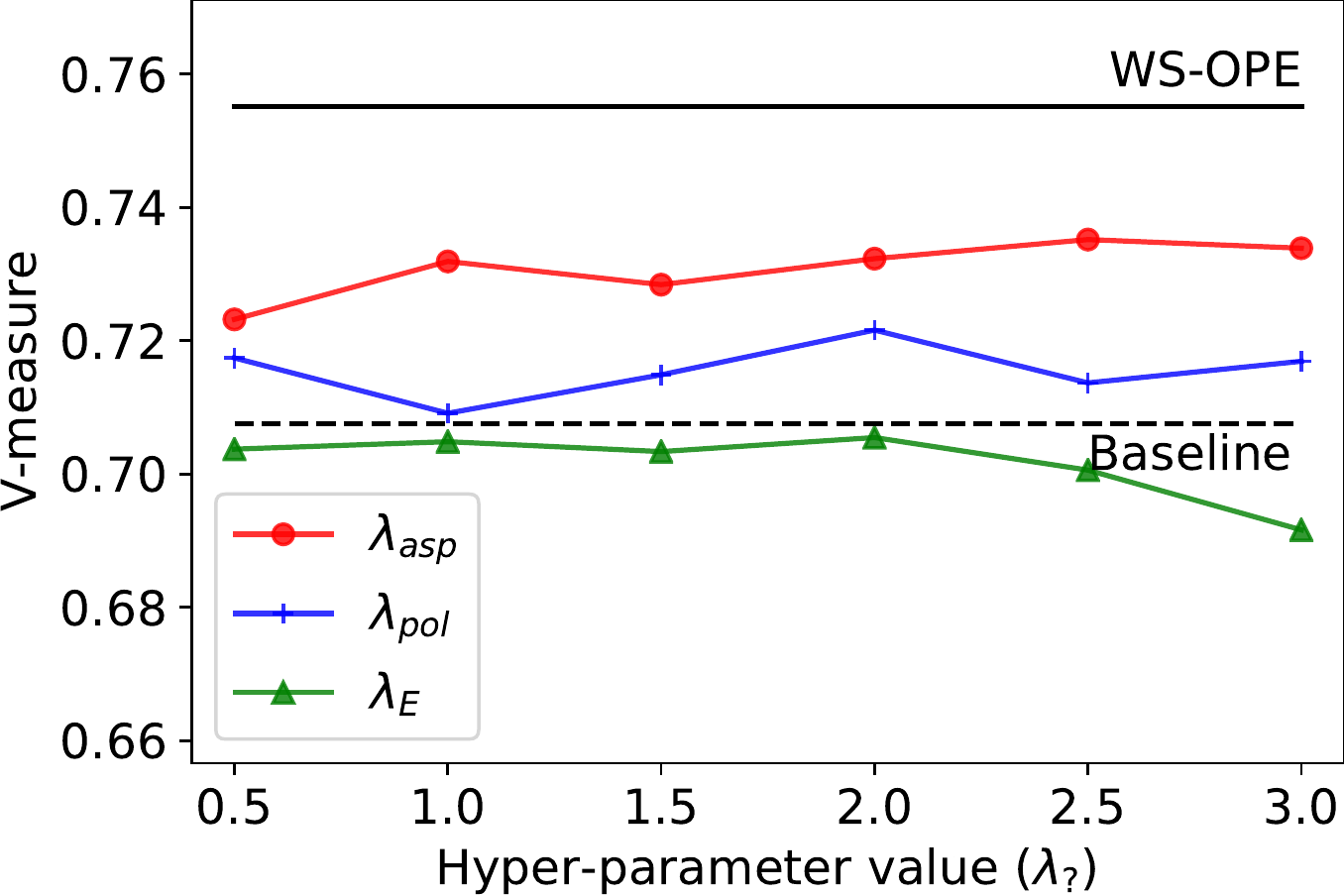}
\endminipage\hfill
\endminipage
\caption{Sensitivity analysis on hyper-parameters $\lambda_{asp}$, $\lambda_{pol}$, and $\lambda_{E}$. The solid line is the performance of \canonical{} and the dotted line is that of the best-performing baseline model (from Table~\ref{fig:cluster:all}).}\label{fig:sensitivity}
\vspace{-4mm}
\end{figure*}

\subsubsection{Methods}
To understand the benefits of our learned opinion phrase embeddings, we evaluate whether they can consistently improve the performance of existing clustering algorithms. Here we select three representative clustering algorithms, $k$-means, Gaussian Mixture Models (GMM)~\cite{Bishop:2006:PRML}, and Correlation Clustering over similarity score~\cite{Bansal:2004:CorrelationClustering,Elsner-schudy-2009-bounding:CorrelationClustering}. For $k$-means and GMM, we set $k=50$ and $k=20$ for \hotel\ and \restaurant\ datasets, respectively; we set $\theta=0.85$ for Correlation Clustering for both datasets. We compared the following methods, which use the same word embedding model ({\tt glove.6B.300d}):

\noindent {\bf \glove:} 
We first calculate the average word embeddings for opinion term and aspect term using GloVe, and then concatenate the aggregated average embedding as the final opinion phrase embedding.

\noindent {\bf ABAE:} We fine-tune the word embedding model without additional labels (i.e., $\lambda_\mathtt{asp}=\lambda_\mathtt{pol}=\lambda_E=0$), which can be considered an ABAE model~\cite{he2017unsupervised}.

\noindent {\bf \canonical}: We learn opinion phrase embeddings based on GloVe word embeddings with weak-supervision from aspect category, polarity, and explanation with the following hyper-parameters: $\lambda_{asp}=1.0, \lambda_{pol}=0.5, \lambda_{E}=0.2$. We obtained the explanations by our explanation classifier (Section~\ref{sec:exp})\footnote{We use the same trained model for both \hotel\ and \restaurant.}.

\subsubsection{Result analysis}
As shown in Table~\ref{fig:cluster:all}, 
our learned opinion phrase representations (\canonical) achieve the best performance among all settings and consistently boost the performance of existing clustering algorithms compared to the baseline methods in both \hotel\ and \restaurant\ domains. In addition, we confirm that our model significantly benefits from the additional weak supervision from opinion and explanation mining as our method significantly improves the performance compared to ABAE, which does not use the weak supervision.

\subsubsection{Usefulness of mined explanations}\label{subsubsec:usefulness}
To verify if the explanations mined in the previous step contribute to the performance of opinion phrase canonicalization, we conducted an ablation study. We evaluated our method without the intra-cluster loss (i.e., $\lambda_E = 0$), so the learned opinion phrase representations do not consider any explanation relationships between opinion phrases.
Figure~\ref{fig:merge:ablation} shows that the performance of our model degrades without the intra-cluster loss (i.e., mined explanations) but is still significantly better than the baseline ABAE model. The results also confirm that the intra-cluster loss based on mined explanations can boost the performance of opinion phrase canonicalization.

\subsubsection{Hyper-parameter sensitivity} \label{sec:cluster:sensitivity}

We also conducted the sensitivity analysis on the hyper-parameters $\lambda_{asp}$, $\lambda_{pol}$, and $\lambda_E$, which balance the multiple loss functions in Eq.~(\ref{eq:all_loss}), to evaluate the robustness of our model with respect to those hyper-parameters. Specifically, we evaluated our model with different $\lambda_?$ ($? = \{asp, pol, E\}$) $\in$ \{0.5, 1.0, 1.5, 2.0, 2.5, 3.0\} while fixing the other two hyper-parameters as 0. Therefore, we can test the contribution of each loss function with different weights when combined with the base reconstruction loss.

Figure~\ref{fig:sensitivity} shows the results for three evaluation metrics (Homogeneity, Completeness, and V-measure). From the results, we confirm that by using either the aspect category loss or the polarity loss, our model consistently outperforms the baseline models. 
Although only using the intra-cluster loss ($\lambda_E > 0$) does not outperform the baseline, we have shown the usefulness of the intra-cluster loss when combined with the other loss functions in \ref{subsubsec:usefulness}.

\begin{figure*}[t]%
    \subfloat[Opinion phrase embeddings of AvgWE (i.e., before representation learning)]{{\includegraphics[width=.46\linewidth]{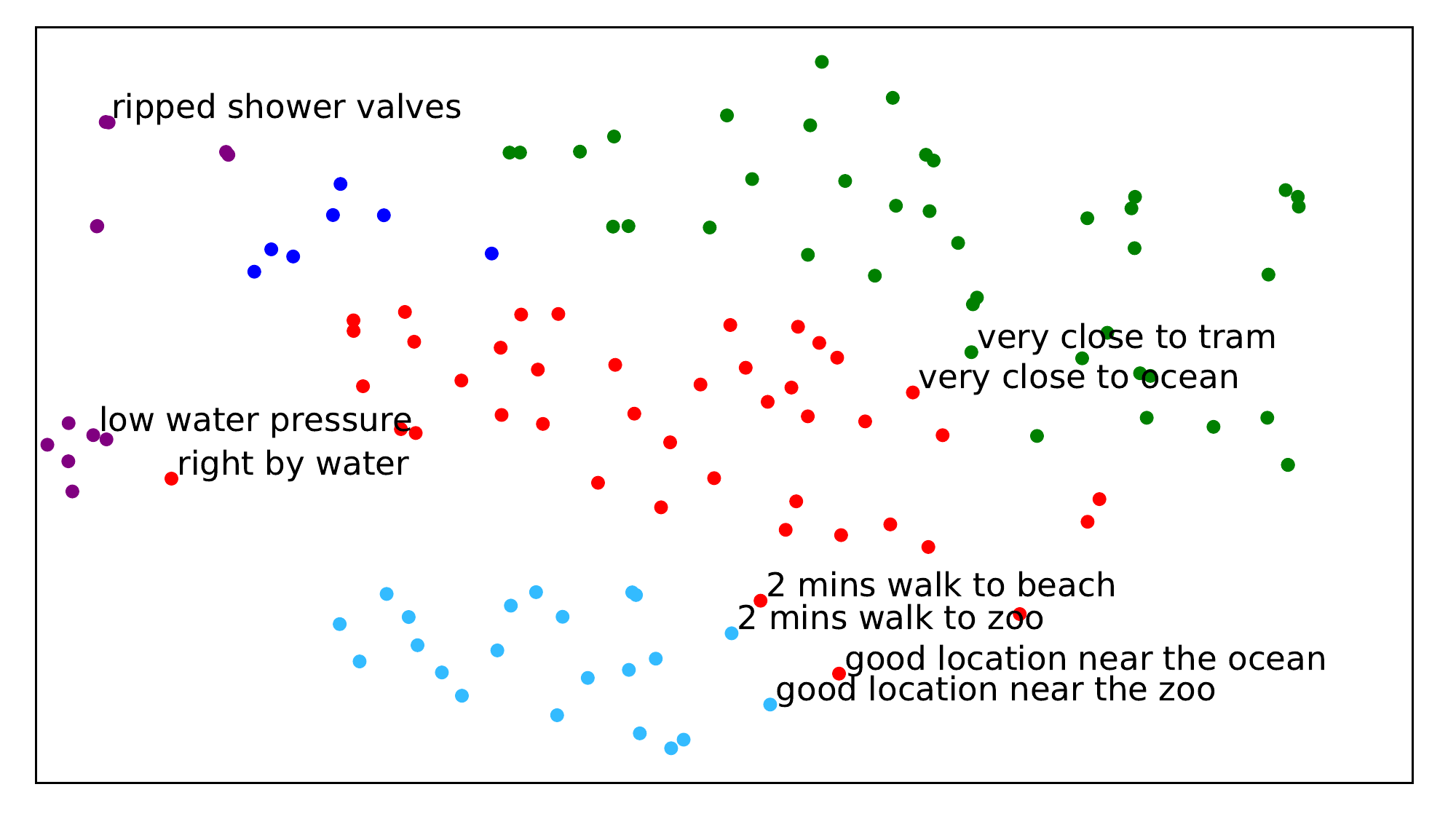} }} \label{fig:emb:before}
    \hfill
    \subfloat[Opinion phrase embeddings of \canonical{} (i.e., after representation learning)]{{\includegraphics[width=.46\linewidth]{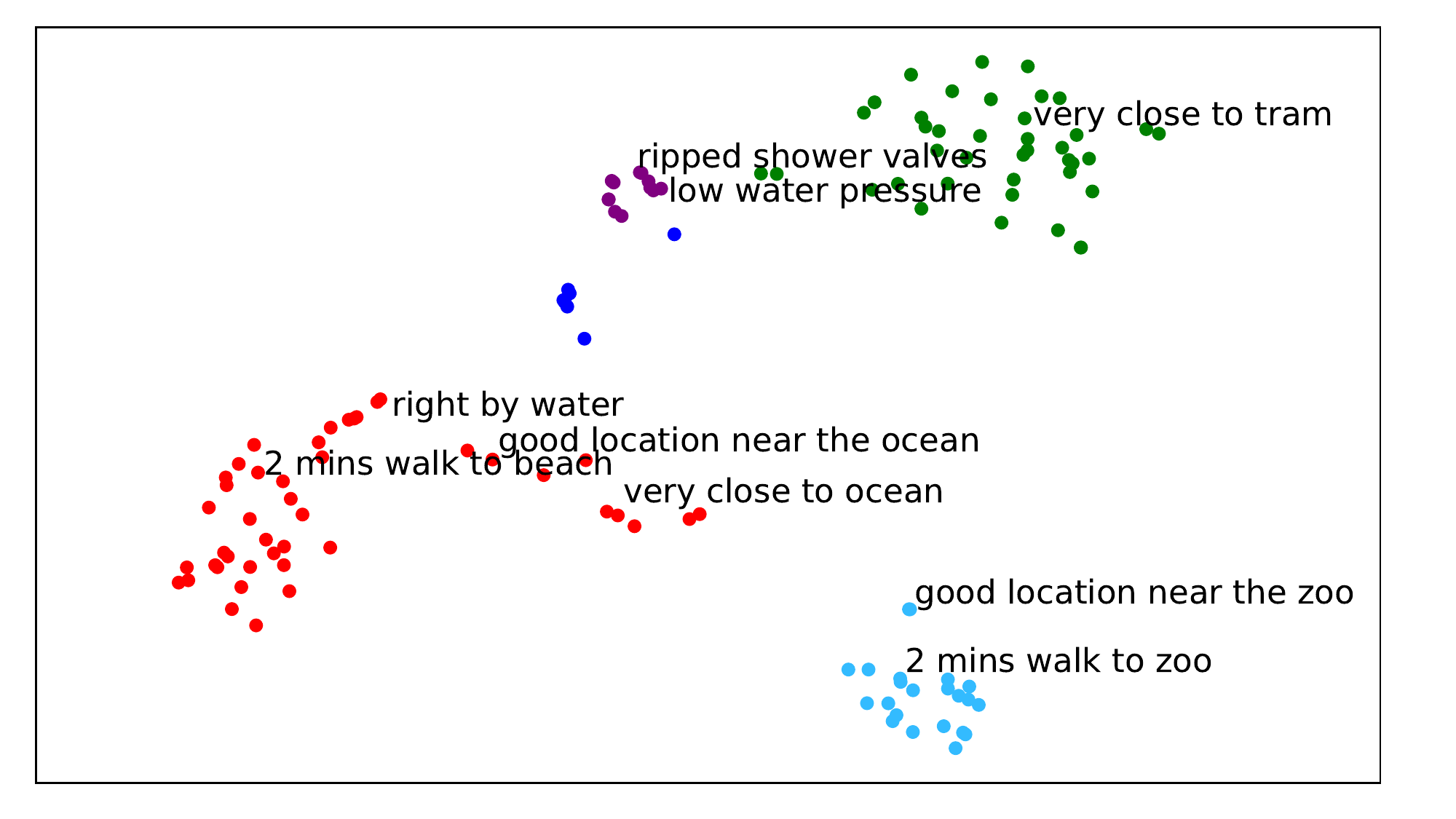} }}%
    \caption{Embedding space comparison. Color-coding denotes true cluster assignments. After learning representations, semantically similar opinion phrases are significantly closer to each other and irrelevant ones are further apart in the embedding space, allowing easier opinion phrase canonicalization.}%
    \label{fig:emb}%
    \vspace{-4mm}
\end{figure*}

\subsubsection{Embedding space visualization}
We also present a qualitative analysis of how our \canonical{} helps to canonicalize opinion phrases. Figure~\ref{fig:emb} shows a two-dimensional t-SNE projection~\cite{maaten2008visualizing} of embeddings for a fraction of the opinion phrases about a hotel. The opinion phrase embeddings obtained before and after representation learning are shown on the left and right side of the figure, respectively. The color codes denote true cluster assignments. 

We observe that the original vectors appear more uniformly dispersed in the embedding space, and hence, the cluster boundaries are less prominent. Additionally, we annotated the figure with a number of particularly problematic cases. For example, each of the following pairs of opinion phrases (\asop{``very close to tram''}, \asop{``very close to ocean''}), (\asop{``2 mins walk to beach''}, \asop{``2 mins walk to zoo''}), and (\asop{``good location near the ocean''}, \asop{``good location near the zoo''}), appear in close proximity when they should belong to different clusters. After learning the representations, the problematic pairs of vectors are now clearly separated in their respective clusters.

\begin{figure}[!t]
    \centering
    \includegraphics[width=0.4\textwidth]{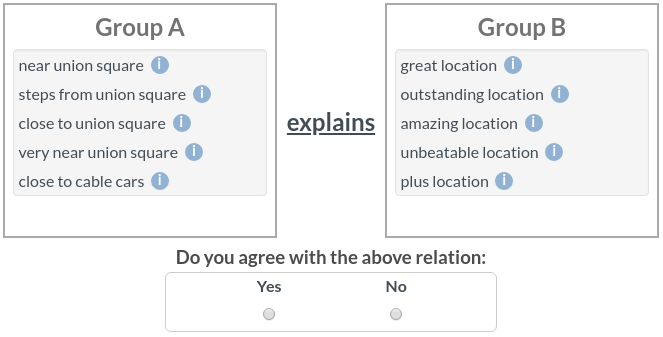}
    \caption{Example question for our user study. Full reviews can be seen when hovering over the (i) icon for each opinion phrase.}
    \label{fig:study}
    \vspace{-3mm}
\end{figure}

\subsection{Opinion Graph Quality: User Study}
In addition to the automatic evaluation of the explanation mining and opinion phrase canonicalizing modules, we designed a user study to verify the quality of the final opinion graphs produced by \system. Assessing the quality of an entire opinion graph at once is impractical due to its size and complexity. Instead, we broke down the evaluation of each generated graph into a series of pairwise tests, where human judges were asked to verify the explanation relation (or lack thereof) between pairs of nodes in the graph. 

More specifically, given a predicted graph $G = (N, E)$ about an entity, we sampled node pairs $(n_i, n_j)$, so that we get a balanced number of pairs for which we predicted the existence or absence of an explanation relation. For every pair, we present the two nodes to the user and show five member opinion phrases from each one. We further show the predicted relation between the nodes (\textsl{``explains"} or \textsl{``does not explain"}) and ask the users if they agree with it. 
An example is shown in Figure~\ref{fig:study}. 

We generated examples for 10 hotels (i.e., their constructed opinion graphs), amounting to 166 node pairs (or questions) in total. This user study was done via Appen's highest accuracy (Level-3) contributors, who obtained no less than 80\% accuracy on our test questions. Every question was shown to 3 judges and we obtained a final judgment for it using a majority vote\footnote{The inter-‐annotator agreement between contributors is 85.6\%.}. The judges \textsl{agreed} with our predicted relation in 77.1\% of cases.

\subsection{Opinion Graph Usefulness}
To evaluate the usefulness of the generated summaries, we further presented the predicted clusters and explanation relations to crowd workers and let them judge the usefulness of such produced information. Note that our clusters of near-synonym opinions and the explanation relations are largely absent from existing travel/service booking websites. We again used Appen's highest accuracy (Level-3) contributors for this task. We presented $164$ clusters-explanation pairs to $1006$ workers and observed that $80.12\%$ of workers found our clusters and explanations useful. 

Furthermore, based on \system{}, we built a summary explorer that visualizes the generated opinion graphs~\cite{wang2020extremereader}. By leveraging the provenance of extracted opinion phrases and explanations, this summary explorer also allows users to access the \textit{``evidence''} (original reviews) of the nodes/edges, thus making \system{} more transparent and interpretable to end-users. Lastly, with the help of an additional abstractive summarization system~\cite{opiniondigest}, \system{} can further generate textual summaries.
\section{Related Work}\label{sec:related}
\noindent
{\bf Opinion Mining:} There has been work of mining opinions from online reviews since \cite{hu2004kdd, hu2004aaai, qiu2011coling, liu2012sentiment, pontiki2015semeval, pontiki2016semeval, xu2019bert}. Those studies developed opinion extraction systems using association mining techniques to extract frequent noun phrases from reviews and then aggregates sentiment polarity scores. These form the aspect-based opinions of online product reviews. 
Opinion Observer~\cite{Liu:2005:OpinionObserver} extended the method to build a system that visualizes the polarity information of each aspect with bar plots. 
\system{} goes one step further from existing opinion mining techniques. Based on extracted opinions, it organizes opinions into an opinion graph such that it includes (a) explanation relationships between opinions and (b) canonicalizes opinions to exclude redundancy. 

\noindent
{\bf Explanation Classifier:} Recognizing Textual Entailment (RTE)~\cite{Dagan:2005:PascalRTE} and Natural Language Inference (NLI)~\cite{snli:emnlp2015} are the tasks to judge if given two statements, whether one statement can be inferred from the other. 
These tasks are usually formulated as a sentence-pair classification problem where the input is two sentences.
A major difference between RTE models and our explanation classifier is that our classifier judges if an opinion phrase explains another opinion phrase in the same review text.  The two opinion phrases may appear in the same sentence or may appear in different sentences. Hence, as described in Section \ref{sec:exp}, we added another task, i.e., explanation existence judgment, in addition to the opinion-phrase classification task to improve the performance using a multi-task learning framework. The goal of relation classification~\cite{zhou2016attention, lin2016neural, wu2019enriching} is to classify the relationship between a pair of entities. For example, determining the relation between entity ``Ms. Ruhl'' and entity ``Chicago'' given a context sequence ``\underline{Ms. Ruhl}, 32, grew up in suburban \underline{Chicago}.''. Similar to the explanation mining problem, the input of relation classification also includes both the context sequence and a pair of entities. However, different from \expcls, existing relation classification models do not incorporate word-by-word alignments between entities. This is because for relation classification, such alignment (e.g., alignment between ``Ms. Ruhl'' and ``Chicago'') is not very useful compared to the context sequence that connects the given pair of entities (e.g., ``grew up in suburban''). 

\noindent
{\bf Opinion Phrase Canonicalization:}
Aspect-based auto-encoder~\cite{he2017unsupervised} is an unsupervised neural model that clusters sentences while learning better word embeddings for sentiment analysis. It showed better performance than conventional topic models (e.g., LDA~\cite{Blei:2003:LDA}, Biterm Topic Model~\cite{Yan:2013:BTM}) in aspect identification tasks. 
Our \canonical{} extends their approach by (1) having an opinion phrase encoder that consists of two encoders for aspect and opinion terms, and (2) incorporating additional sentiment signals such as sentiment polarity and aspect category in a weakly-supervised manner.

Opinion phrase canonicalization is closely related to KB canonicalization~\cite{Galarraga:2014:CIKM:Canonicalizing,Vahishth:2018:CESI}, which canonicalizes entities or relations (or both) by merging triples consisting of two entities and a relation, based on the similarity. Gal\'{a}rraga et al.~\cite{Galarraga:2014:CIKM:Canonicalizing} proposed several manually-crafted features\footnote{\cite{Vahishth:2018:CESI} showed that word-embedding features using GloVe outperformed the methods in \cite{Galarraga:2014:CIKM:Canonicalizing}. Thus, we consider GloVe word embeddings as a baseline for the opinion phrase canonicalization task.} for clustering triples.
CESI~\cite{Vahishth:2018:CESI} uses side information (e.g., entity linking, WordNet) to train better embedding representations for KB canonicalization.
The difference from KB canonicalization is that \system{} does not rely on external structured knowledge such as WordNet or KBs, which were used for those models. This is mainly because it is not straightforward to construct a single KB that reflects a wide variety of subjective opinions written in reviews. Instead, we aim to construct an opinion graph for each entity.

\section{Conclusion} \label{sec:conclusion}
We present \system, a system that extracts opinions and constructs an explainable opinion graph from reviews.

\system{} consists of four components including a novel explanation mining component and an opinion phrase canonicalization component, which we developed to construct opinion graphs from online reviews.
Our experimental results  show that our methods significantly perform better than baseline methods in both the task of classifying explanations and canonicalizing opinion phrases by up to $5.4\%$ and, respectively, $12.2\%$.
In addition, our user study confirmed that human judges agree with the explanation relationships depicted in our opinion graph in more than 77\% of the cases.
We created labeled datasets for explanation mining and opinion phrase canonicalization tasks and we made these datasets publicly available for future research.

\bibliographystyle{ACM-Reference-Format}
\bibliography{main}

\end{document}